\pdfoutput=1
\documentclass{article}
\makeatletter
\def\input@path{{./}{drafts/alignment_notes/}}
\makeatother

\usepackage[preprint]{neurips_2026}

\usepackage{amsmath,amssymb,amsthm}
\usepackage{graphicx}
\usepackage{booktabs}
\usepackage{array}
\PassOptionsToPackage{bookmarks=false}{hyperref}
\usepackage{hyperref}
\usepackage{bm}
\usepackage{microtype}
\usepackage{xcolor}
\usepackage{subcaption}
\hypersetup{hidelinks,breaklinks=true}

\graphicspath{{./}{drafts/alignment_notes/}}

\newtheorem{lemma}{Lemma}
\newtheorem{proposition}{Proposition}

\newif\ifpaperincludechecklist
\paperincludechecklisttrue

\newcommand{\RQ}{\mathrm{RQ}}
\newcommand{\Var}{\mathrm{Var}}
\newcommand{\Red}{\mathrm{Red}}     
\newcommand{\Syn}{\mathrm{Syn}}
\newcommand{\ITY}{I(T;Y)}

\newcommand{\MainTextFigurePacing}{%
  \setcounter{topnumber}{1}%
  \setcounter{bottomnumber}{1}%
  \setcounter{dbltopnumber}{1}%
  \setcounter{totalnumber}{2}%
  \setlength{\textfloatsep}{8pt plus 1pt minus 2pt}%
  \setlength{\floatsep}{8pt plus 1pt minus 2pt}%
  \setlength{\intextsep}{8pt plus 1pt minus 2pt}%
  \renewcommand{\topfraction}{0.90}%
  \renewcommand{\dbltopfraction}{0.90}%
  \renewcommand{\bottomfraction}{0.70}%
  \renewcommand{\textfraction}{0.03}%
  \renewcommand{\floatpagefraction}{0.75}%
  \renewcommand{\dblfloatpagefraction}{0.75}%
}

\newcommand{\AppendixFigurePacing}{%
  \setcounter{topnumber}{2}%
  \setcounter{bottomnumber}{1}%
  \setcounter{dbltopnumber}{2}%
  \setcounter{totalnumber}{4}%
  \renewcommand{\topfraction}{0.92}%
  \renewcommand{\dbltopfraction}{0.92}%
  \renewcommand{\bottomfraction}{0.75}%
  \renewcommand{\textfraction}{0.04}%
  \renewcommand{\floatpagefraction}{0.70}%
  \renewcommand{\dblfloatpagefraction}{0.70}%
}

\paperincludechecklistfalse

\title{Task Relevance Is Not Local Replaceability: A Two-Axis View of Channel Information}

\author{%
  \footnotesize Houman Safaai\textsuperscript{1}\thanks{Correspondence: \texttt{houman\_safaai@harvard.edu}.},
  Andrew T. Landau\textsuperscript{2},
  Celia C. Beron\textsuperscript{1,2},
  Yasin Mazloumi\textsuperscript{1},
  Bernardo L. Sabatini\textsuperscript{1,2} \\
  \normalfont\textsuperscript{1}Kempner Institute for the Study of Natural and Artificial Intelligence \\
  \normalfont Harvard University, Cambridge, MA 02138, USA \\
  \normalfont\textsuperscript{2}Department of Neurobiology, Howard Hughes Medical Institute \\
  \normalfont Harvard Medical School, Boston, MA 02115, USA
}

\begin{document}
\maketitle

\begin{abstract}
Channel importance in vision networks is usually summarized by a single score.
That summary hides two different questions: how much a channel is related to the task, and whether its function can be supplied by same-layer peers when the channel is removed.
We call the second property local replaceability.
We introduce a two-axis view that separates these questions.
The local axis measures input capture and peer overlap, while the target axis measures task information and target-excess information.
Across ResNet-18, VGG-16, and MobileNetV2 trained on CIFAR-100, the two axes are weakly aligned, induce different channel groupings, and separate rapidly during training despite being strongly coupled at random initialization.
A Gaussian linear analysis accounts for how this separation can arise through residualized gradient directions, and lesion plus peer-replacement experiments show that peer support refines removability beyond input capture and task relevance alone.
Under the fixed FLOPs-matched pruning protocol, local-axis metrics are more reliable predictors of removability than target-axis metrics across the three CIFAR-100 backbones, with the same direction preserved in stress tests on CIFAR-10, Tiny-ImageNet, ImageNet-100, and a ConvNeXt-T/ImageNet-100 pilot.
These findings identify an axis-level distinction rather than a universal ranking of pruning scores: local replaceability is a more reliable guide to removability than target relevance, while norm-based baselines remain competitive in architectures such as VGG-16.
Relevance-based scores ask what a channel says about the task; pruning asks whether the network still needs that channel when its peers remain available.
\end{abstract}

\MainTextFigurePacing

\section{Introduction}
\label{sec:intro}

Channel-importance scores are often treated as if they measure one property.
In practice, they mix two different questions.
The first is target relevance: what does a channel say about the task?
The second is local replaceability: if that channel is removed, can the remaining channels in the same layer supply the information needed for the network to function?
These questions need not agree.
A highly predictive channel may be removable with little loss when several peers carry overlapping local information, whereas a less predictive channel may be costly to remove when no peer covers its role.
For interpretation, pruning, and regularization, channel importance should therefore be described not only by what a channel contributes to the task, but also by whether its function is locally replaceable.

This distinction is easy to miss because many successful channel scores compress several effects into a single scalar.
Magnitude, gradient sensitivity, reconstruction error, geometric-median redundancy, and channel-independence scores each capture useful pruning signals~\NoHyper\citep{li2017pruning,molchanov2019importance,luo2017thinet,he2019filter,sui2021chip}\endNoHyper, but they do not explicitly distinguish task relevance from recoverability by same-layer peers.
Information-bottleneck analyses summarize representations through compression--prediction tradeoffs~\NoHyper\citep{tishby2000information,shwartzziv2017opening,saxe2018information,goldfeld2019estimating}\endNoHyper, and feature-selection methods distinguish relevance from redundancy for input variables relative to an external target~\NoHyper\citep{peng2005mrmr,westphal2025pidf}\endNoHyper.
Here we ask a related but different question: within a trained internal layer, can the function of one learned channel be recovered from the other channels in that same layer?

\begin{figure}[tbp]
    \centering
    \includegraphics[width=0.98\columnwidth]{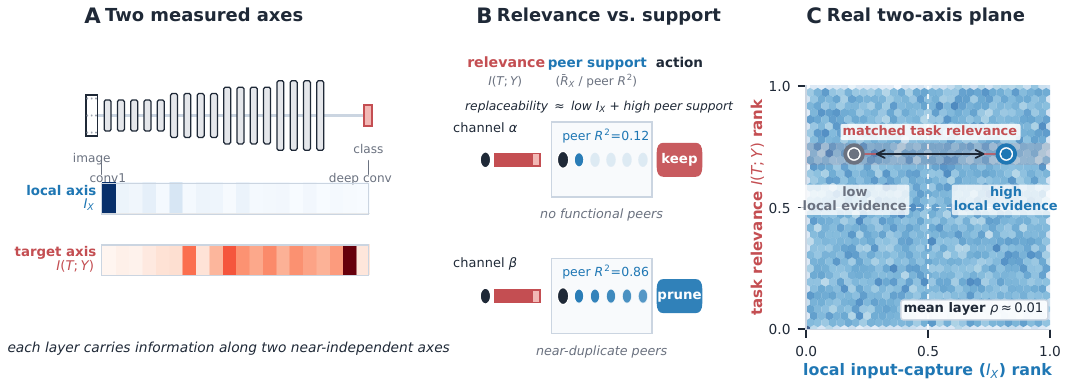}
    \caption{\textbf{Conceptual overview.}
    (A) Local input variation and target relevance need not follow the same depth profile.
    (B) Channels with similar task relevance can differ in peer support, so relevance alone does not determine removability.
    (C) Real channels occupy a weakly coupled two-axis plane; the highlighted band holds $I(T;Y)$ fixed while local input capture $I_X$ varies.}
    \label{fig:schematic}
\end{figure}

We formalize this question through a two-axis view of channel information, illustrated in Figure~\ref{fig:schematic}.
The local axis measures input capture $I_X(i)$ and overlap with same-layer peers $\bar R_X(i)$.
A channel with high input capture and low peer overlap carries distinctive local structure and is therefore harder to replace.
The target axis measures task information $I(T;Y_i)$ and a pairwise target-excess proxy $\Syn_i$.
The axes are asymmetric by design: the target side decomposes information about a task, whereas the local side measures throughput and peer recoverability within a layer.

Importantly, this separation is not automatic.
At random initialization, local input capture and task relevance are strongly aligned.
During training, however, their alignment drops sharply, and final CIFAR-100 models organize channels differently along the two axes.
Within-layer $I_X$ and $I(T;Y)$ are nearly uncorrelated, local and target clusterings remain close to the permutation null, and target-side redundancy under the scalable Gaussian-MMI proxy collapses to singleton task relevance rather than same-layer peer overlap.
Non-parametric BROJA checks preserve the distinction without relying on the MMI proxy.
A representative ResNet-18 layer makes the practical consequence visible: about 26\% of channels selected by a one-dimensional ranking differ from those selected by a two-axis score (Appendix~\ref*{app:teaser}).

The rest of the paper supports this view in three steps: structural separation in trained CNNs, a Gaussian residualization mechanism for learned decoupling, and intervention tests through lesions and FLOPs-matched pruning.
The axes also leave a weak but measurable trace across layers through weight connectivity.

Together, these results support an axis-level conclusion rather than a universal ranking of pruning methods.
Pruning is primarily a replaceability intervention: it asks whether a channel is still needed when the rest of the layer remains available.
Task relevance remains important for understanding what a channel represents, but it is not enough to predict whether that channel can be removed.
This qualification also explains why norm-based scores remain competitive in some architectures.
In VGG-16, for example, magnitude remains the strongest scalar baseline, showing that architecture and allocation details still matter even when the axis-level local-over-target pattern holds.
Appendix~\ref{app:rule_experiments} reports two pilot extensions (a local-axis decorrelation regularizer and a frozen-ViT transfer test).

\section{Two Axes of Channel Information}
\label{sec:two_axes}

\subsection{Setup and metrics}
\label{sec:metrics}

We next turn the conceptual distinction into post-hoc channel diagnostics computed on frozen networks with a fixed calibration set.
The metrics separate what a channel contributes to the task from how easily its function can be recovered by peers, without introducing a new training objective or pruning rule.
We use ResNet-18, VGG-16 with BatchNorm, and MobileNetV2 trained on CIFAR-100 with five seeds each~\NoHyper\citep{he2016deep,simonyan2015very,sandler2018mobilenetv2}\endNoHyper.
For every trained model, we compute channel metrics on the same fixed 5000-image calibration subset of the training split.
For layer $\ell$ with $N_\ell$ channels, $Y_i$ denotes the stored Conv2d forward-hook activation: post-convolution, pre-BatchNorm, pre-ReLU, and pre-residual-addition unless a post-BN control is explicitly marked.
Target-axis quantities use global-average-pooled activations, so each image contributes one value per channel.
Local overlap uses the full same-layer channel correlation matrix, with spatial positions treated as additional samples for convolutional activations.
For the input-capture proxy, convolutional inputs are unfolded into patches before estimating the pre-activation covariance $\Sigma_X$.

For channel $i$ with weight vector $w_i$, pre-activation input covariance $\Sigma_X$, and output $Y_i$, the local axis is represented by two operational quantities:
\begin{align}
    \text{Input-capture proxy:}\quad I_X(i) &\equiv \tfrac{1}{2}\log\!\bigl(1 + s_i / \sigma_0^2\bigr), \qquad s_i = w_i^\top \Sigma_X w_i = \RQ_i \|w_i\|^2, \label{eq:ixy}\\
    \text{Within-layer shared information:}\quad \bar{R}_X(i) &= \frac{1}{N_\ell-1}\sum_{j \neq i} \bigl[-\tfrac{1}{2}\log(1 - \rho_{ij}^2)\bigr]. \label{eq:red}
\end{align}
All information quantities are reported in nats.
Here $\RQ_i = w_i^\top \Sigma_X w_i / \|w_i\|^2$ is the Rayleigh quotient, $\sigma_0^2$ is the fixed Gaussian noise variance used by the input-information proxy, and $\rho_{ij}$ is the Pearson correlation between same-layer channels $i$ and $j$.
In the main analyses, $\sigma_0^2$ is set separately in each layer to the median positive signal power, $\mathrm{median}_i(s_i)$, with implementation details and sensitivity checks reported in Appendix~\ref{app:experiments}.
In practice, the saved $I_X$ scores are calibrated Gaussian signal-power proxies rather than literal mutual-information estimates for deterministic continuous channels; because the noise floor is set per layer, $I_X$ is primarily used as a within-layer coordinate and ranking.
$I_X(i)$ measures how strongly a channel captures or transmits input variation, while $\bar{R}_X(i)$ measures how much of that signal overlaps with peers.
Equation~\ref{eq:red} is therefore the per-channel peer-overlap coordinate used throughout the figures.
Together, they characterize \emph{replaceability}: a channel with high $I_X$ and low $\bar{R}_X$ carries distinctive, hard-to-replace local structure.
Unlike weight magnitude, $\bar R_X$ is activation-derived peer overlap; it is the replaceability component that norm-like scores cannot express, even when $I_X$ behaves similarly to magnitude in architectures such as VGG-16.
For BatchNorm networks, these raw convolutional-stage quantities are representation diagnostics rather than parameterization-invariant saliency scores; Appendix Table~\ref{tab:bn_reparam_control} reports the Conv--BN reparameterization control.

The target axis uses the logit margin $T = f_z(x) - \max_{c \neq z} f_c(x)$ as a continuous task variable, where $z$ is the ground-truth class:
\begin{align}
    \text{Task MI:}\quad I(T; Y_i) &= -\tfrac{1}{2}\log(1 - \rho_{T,i}^2), \label{eq:taskmi}\\
    \text{Target redundancy:}\quad \Red_T(i) &= \tfrac{1}{m_i}\sum_{j \in \mathrm{Top}\text{-}m(i)} \min\{I(T;Y_i),I(T;Y_j)\}, \label{eq:redt}\\
    \text{Target-excess proxy:}\quad \Syn_i &= \tfrac{1}{m_i}\sum_{j \in \mathrm{Top}\text{-}m(i)} \Syn(T; Y_i, Y_j), \label{eq:syn}
\end{align}
where $\rho_{T,i}$ is the correlation between $T$ and the pooled activation $Y_i$, $\mathrm{Top}\text{-}m(i)$ denotes the $m_i=\min(10,N_\ell-1)$ channels with the largest singleton task MI among $j\neq i$, and $\Syn(T; Y_i, Y_j) = I(T; [Y_i, Y_j]) - \max\{I(T;Y_i), I(T;Y_j)\}$ is the pairwise Gaussian-MMI target-excess proxy: additional task information from the pair beyond the better singleton.
The per-channel target-redundancy score in Eq.~\ref{eq:redt} is the $R_T$ or $\Red_T$ quantity shown in the figures.
The ground-truth margin is the main target variable, but the separation is not tied to this choice: Appendix Table~\ref{tab:target_sensitivity_control} repeats the analysis for predicted margins, class logits, negative loss, and one-vs-rest label targets.
The target-excess coordinate is also stable to the partner-selection rule: varying $m$ and selecting partners by singleton task MI, joint MI, direct excess, or random task-relevant peers preserves weak local-axis alignment (Appendix Table~\ref{tab:synergy_partner_control}).

The two axes have different mathematical status.
The target axis admits a PID decomposition about $T$~\citep{williams2010nonnegative}, whereas the local axis has no single target to decompose about; it combines the throughput proxy $I_X$ with the peer-overlap matrix $\bar R_X$.
This asymmetry avoids treating deterministic continuous activations as literal finite mutual-information estimates without an explicit noise model.

All main CIFAR-100 structural analyses use the three-backbone, five-seed suite above; pruning details and recovery-budget controls are in Appendix~\ref{app:experiments}.

\begin{figure*}[tbp]
    \centering
    \includegraphics[width=0.98\textwidth]{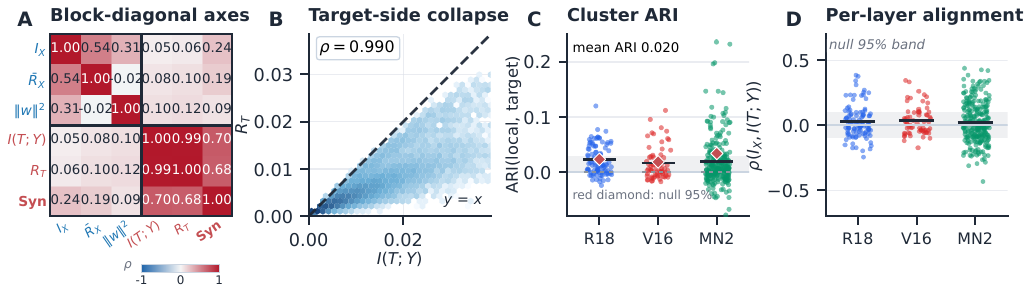}
    \caption{\textbf{Weakly aligned axes of channel information (CIFAR-100; 3 architectures $\times$ 5 seeds, all layers pooled).}
    (A) Spearman rank correlation matrix after within-layer rank normalization for six channel metrics: three local-axis ($I_X$, $\bar{R}_X$, $\|w\|^2$) and three target-axis ($I(T;Y)$, $\Red_T$, Syn).
    The clear block-diagonal structure (high within-block, near-zero between-block) supports weak cross-axis alignment rather than a single collapsed importance dimension.
    The partition line separates the two axes.
    (B) MMI target-side sanity check: target-directed redundancy $\Red_T$ lies almost exactly on the $y=x$ line against task MI $I(T;Y)$ (pooled $\rho=0.990$ in the panel; per-layer summaries $\rho\geq0.994$), showing that Gaussian-MMI target redundancy is determined by singleton task relevance under this proxy (Lemma~\ref{lem:mmi_collapse}).
    (C) Adjusted Rand Index between local-axis and target-axis clusterings, shown against the permutation null.
    Points show observed layer$\times$seed ARIs, black bars mark model means, gray ticks mark null means, and red diamonds mark the null 95th percentile.
    The observed means remain near the null threshold (overall mean $= 0.020$), showing that the two axes define different groupings.
    (D) Per-layer cross-axis correlations remain centered near zero for all three models; the shaded band marks the near-null region used as the weak-alignment reference.
    Empirical BROJA/KSG target-side checks are reported in Appendix~\ref{app:controls} and Appendix~\ref{app:theory_figure}.}
    \label{fig:two_axes}
\end{figure*}

\subsection{Weak alignment of the two axes}
\label{sec:independence}

The first question is whether the two axes capture different structure or simply two noisy views of one ranking.
In trained networks they do not collapse: in the figure-level layer summary over the canonical CIFAR-100 suite, the within-layer Spearman correlation between $I_X$ and $I(T;Y)$ is $\bar{\rho} = 0.012 \pm 0.08$ (mean $\pm$ SD over layer--seed cells).
The Adjusted Rand Index between clusterings based on local features $(I_X, \bar{R}_X)$ and target features $(I(T;Y), \Syn)$ is similarly close to the permutation null in the same summary: $\overline{\mathrm{ARI}} = 0.020$, with $p > 0.05$ for 95\% of layers.
The conclusion is unchanged under model-seed bootstrap and leave-one-architecture-out aggregation: the model-seed bootstrap gives $\rho=-0.016$ with 95\% CI $[-0.028,-0.002]$ and ARI $=0.035$ with 95\% CI $[0.029,0.042]$ (Appendix Table~\ref{tab:hierarchical_robustness}).
The difference from the layer--seed headline values reflects aggregation and weighting rather than a different experiment.
Thus, channels that carry large amounts of input-side signal are not necessarily the channels most aligned with the task margin.

\subsection{MMI sanity check: target redundancy is not replaceability}
\label{sec:pid_collapse}

The term ``redundancy'' is overloaded: PID redundancy is information shared about a target, whereas local replaceability is overlap among channels within the same layer.
The Gaussian assumption gives closed-form MI estimates, but the collapse below is a property of the MMI redundancy definition rather than a property of channel replaceability.

\begin{lemma}[MMI target-side redundancy collapse]
\label{lem:mmi_collapse}
Under the MMI redundancy measure~\citep{barrett2015exploration}, the target-directed redundancy is:
\begin{equation}
    \Red_T(T; Y_1, Y_2) = \min\{I(T; Y_1), I(T; Y_2)\}.
\end{equation}
For jointly Gaussian variables, the singleton MI terms have the closed-form correlation estimates used in our scalable proxy; the MMI redundancy itself depends only on those individual task-MI terms, not on the channel-channel correlation $\rho_{12}$.
\end{lemma}

The proof (Appendix~\ref{app:math}) follows directly from the MMI definition; the useful content is what it rules out for scalable MMI-style pruning proxies.
Empirically, the pooled Figure~\ref{fig:two_axes}B correlation is $0.990$, per-layer summaries satisfy $\rho(\ITY,\Red_T)\geq0.994$, and $\rho(\bar{R}_X,\Red_T)\approx0.05$.
Under this scalable Gaussian-MMI proxy, target-side redundancy and unique-information terms are determined by singleton task relevance, while pairwise target-excess information remains a secondary target-side quantity.
The local axis ($I_X$, $\bar{R}_X$) captures a fundamentally different, correlation-dependent structure invisible to the MMI target-redundancy measure.
Appendix~\ref{app:controls} and Appendix~\ref{app:benchmark_consistency} preserve this broad pattern under BROJA/KSG target checks, target-definition controls, and 68 broader runs; ConvNeXt-T is supportive but more coupled.
We therefore use Gaussian MMI as a scalable structural proxy, not as a claim of exact channelwise estimator agreement.

\subsection{Beyond MMI: non-parametric PID robustness}
\label{sec:beyond_mmi}

The MMI collapse clarifies what a scalable Gaussian-MMI target proxy can and cannot measure, but it could still be too tied to the MMI definition.
We therefore repeat the target-side analysis with non-parametric BROJA~\citep{bertschinger2014quantifying} and KSG estimates~\citep{kraskov2004estimating} on the same 5-seed CIFAR-100 runs.
BROJA gives richer target-side atoms than MMI, but its shared information still lands \emph{on the target axis}: pooled across layers, $\rho(\mathrm{SI}, I(T;Y)) \in [0.81, 0.88]$ and $\rho(\mathrm{SI}, \bar R_X) = -0.20$ on ResNet-18 and $-0.18$ on VGG-16.
Thus, shared information about the target and shared information within the layer are distinct quantities.
BROJA SI is target-directed, whereas $\bar R_X$ measures within-layer overlap.
The MMI collapse should therefore be read as a Gaussian-MMI property of the target axis, not as a failure of PID to distinguish the two axes.

The ratio $\mathrm{SI}/\mathrm{CI}$ (shared task-information over target-excess information) follows a U-shaped depth profile in the CIFAR-100 CNN suite: early layers $\approx 0.24$, middle layers $\approx 0.18$ (peak target-excess), and deep layers rising back to $0.25$--$0.37$.
Appendix~\ref{app:triplets} further shows that triplet target-excess over the best pair, $S_3/S_2$, grows from $\sim\!0.2$--$0.3$ near the input to $\sim\!0.4$--$0.55$ near the output.
This suggests middle layers may be harder to compress from the target side because their task-relevant information is distributed synergistically, not shared.

The same separation persists beyond scalars (Appendix~\ref{app:pairwise}): local replaceability shifts from singleton duplicates to distributed hulls with depth, the R-graph is more modular than the target-excess graph, and target-side information increasingly requires triplet-level excess beyond pairs.

\begin{figure*}[tbp]
\centering
\includegraphics[width=0.98\textwidth]{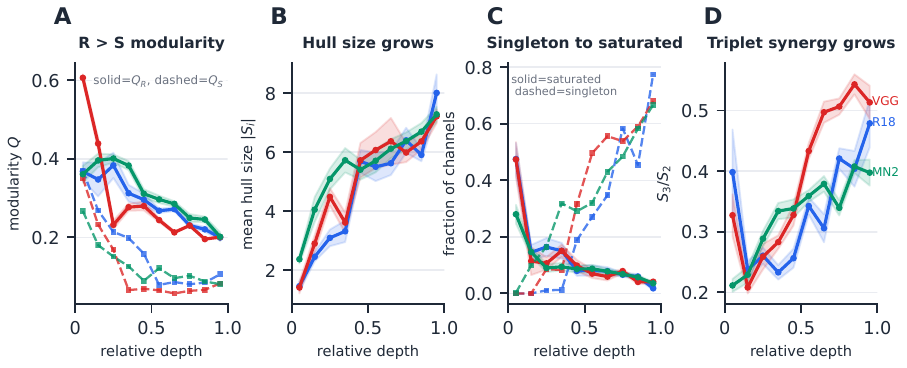}
\caption{\textbf{Higher-order support for the two-axis decomposition.}
(A) Newman modularity of the local redundancy graph ($Q_R$, solid) is larger than that of the target-excess graph ($Q_S$, dashed) at every relative depth.
(B,C) Local replaceability shifts from singleton duplicate regimes to distributed hulls with depth.
(D) Triplet target-excess over the best pair, $S_3/S_2$, rises with depth.
CIFAR-100, 3 backbones $\times$ 5 seeds, mean $\pm$ SEM; R18, VGG, and MN2 denote ResNet-18, VGG-16, and MobileNetV2.}
\label{fig:higher_order_depth}
\end{figure*}

\section{Learning Separates Initially Coupled Axes}
\label{sec:learned}

Trained networks organize channels differently along the local and target axes, but this does not explain where the weak cross-axis alignment comes from.
Training trajectories show that this separation is learned: the axes begin strongly coupled, move together early, and then decouple as local rankings stabilize faster than target rankings.
Gradient geometry explains why this can happen without requiring SGD to optimize a two-axis objective.

\begin{figure*}[t]
    \centering
    \includegraphics[width=0.98\textwidth]{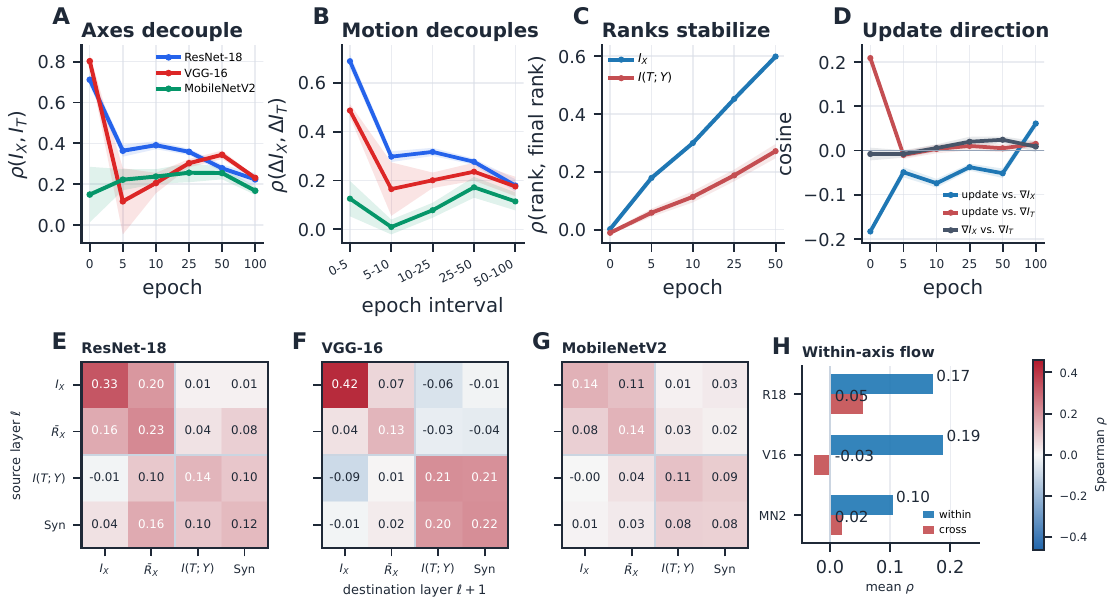}
    \caption{\textbf{The two-axis structure emerges through learning dynamics and propagates weakly across layers.}
    Throughout this figure $I_T \equiv I(T;Y)$ and ``update'' is the SGD update direction $-\nabla L$.
    (A) Cross-axis coupling drops during training.
    (B) Coupled early motion gives way to separated local and target updates; $\Delta I_X$, $\Delta I_T$ are checkpoint-to-checkpoint changes within an epoch interval.
    (C) Local ranks stabilize earlier than target ranks (per-channel rank vs.\ final-checkpoint rank).
    (D) Information-gradient directions ($\nabla I_X$ vs.\ $\nabla I_T$) remain nearly orthogonal; the SGD update is target-aligned mainly at initialization.
    (E--H) Weight-routed source metrics show weak but consistent within-axis flow (Eq.~\ref{eq:weighted_metric}).
    Panels A,B use checkpointed ResNet-18, VGG-16, and MobileNetV2 trajectories; panel C shows ResNet-18, with the same ordering observed on VGG-16 and MobileNetV2; panel D uses 10 ResNet-18 seeds; panels E--H summarize the fixed CIFAR-100 suite.
    R18, V16, and MN2 denote ResNet-18, VGG-16, and MobileNetV2.}
    \label{fig:learned_main}
\end{figure*}

\subsection{Training trajectory: coupled motion gives way to separated axes}

We saved per-epoch metric traces for ResNet-18, VGG-16, and MobileNetV2 from random initialization to convergence on CIFAR-100 (same calibration split and pipeline as the main experiments).
At initialization, the two axes are \emph{strongly coupled}: pooled within-layer Spearman $\rho(I_X, I(T;Y)) = 0.71 \pm 0.03$ on ResNet-18 and $0.72 \pm 0.04$ on VGG-16 across 10 seeds (mean $\pm$ SEM).
Within the first $\sim 10$ epochs, this coupling drops sharply, reaching $0.39 \pm 0.02$ on ResNet-18 and $0.05 \pm 0.07$ on VGG-16 by epoch 10, then ending at $0.22 \pm 0.03$ and $0.02 \pm 0.06$, respectively (Figure~\ref{fig:learned_main}A).
The exact final value depends on checkpoint and aggregation protocol: the checkpointed ResNet-18 trajectory retains moderate positive alignment, whereas the canonical pooled layer--seed analysis over all three backbones remains close to zero.
Because $T$ is the model's own margin and $I(T;Y)$ is correlation-based, the drop in mean task-MI proxy should not be read as a monotone measure of semantic task information.
The important point is that local and target rankings separate rapidly, so weak cross-axis alignment is learned rather than present in random features.
The velocity analysis sharpens this picture.
Early changes along the two axes are coupled: $\rho(\Delta I_X,\Delta I(T;Y))=0.689\pm0.027$ on ResNet-18 and $0.487\pm0.026$ on VGG-16 over epochs 0--5.
By epochs 50--100, the same coupling is much weaker ($0.179\pm0.026$ and $0.175\pm0.040$).
Local ranks also stabilize earlier than target ranks: by epoch 50, rank persistence to the final checkpoint is $0.599\pm0.007$ for $I_X$ versus $0.271\pm0.025$ for $I(T;Y)$ on ResNet-18, with the same ordering on VGG-16 and MobileNetV2 (Figure~\ref{fig:learned_main}C).
A higher-order trajectory control links this maturation to distributed peer structure: the R-vs-S modularity gap is already positive at initialization, while distributed hulls and triplet-irreducible target excess grow during training (Appendix~\ref{app:higher_order_training}).

\subsection{Gradient residualization as a mechanism}

The training trajectories show that the axes separate; a Gaussian channel model explains why this is geometrically plausible.
For Eqs.~\ref{eq:ixy} and~\ref{eq:taskmi}, input capture follows projected input power, whereas task relevance follows a residualized task-covariance direction.
Write $Y_i = w_i^\top X + \varepsilon$ with $\varepsilon \sim \mathcal{N}(0, \sigma_0^2)$ and let $\Sigma_X$ be the pre-activation input covariance.

\begin{proposition}[Gaussian residualization of the two gradients]
\label{prop:grad_orth}
For the Gaussian proxies of Eqs.~\ref{eq:ixy} and~\ref{eq:taskmi}, treat $T$ as an external or stop-gradient scalar target, and let $c=\mathrm{cov}(X,T)$, $b_i=w_i^\top c$, and $s_i=w_i^\top\Sigma_X w_i$.
Up to the scalar factor shown in Appendix~\ref{app:math_grad},
\[
\frac{\partial I_X(i)}{\partial w_i} \;=\; \frac{\Sigma_X w_i}{\sigma_0^2 + w_i^\top \Sigma_X w_i},
\qquad
\frac{\partial I(T; Y_i)}{\partial w_i} \;\propto\; c - \frac{b_i}{\sigma_0^2+s_i}\, \Sigma_X w_i,
\]
so $I_X$ follows the projected signal-power direction whereas, for channels with nonzero singleton task covariance, $I(T;Y_i)$ follows a residualized task-covariance direction.
When $c$ is weakly aligned with the dominant input-power directions, this model provides a simple explanation for the small average cosine observed empirically.
\end{proposition}

The proof is a direct computation in this Gaussian linear model (Appendix~\ref{app:math_grad}); empirically, checkpointed ResNet-18 trajectories keep the two analytic information-gradient directions close to orthogonal throughout training (Figure~\ref{fig:learned_main}D).
The SGD update direction is target-aligned only at initialization ($0.209\pm0.008$ for $\cos(-\nabla L,\nabla I(T;Y))$), and becomes nearly orthogonal to both axes after early training; by the final checkpoint it has only a small local component ($0.061\pm0.009$ for $\cos(-\nabla L,\nabla I_X)$).
This is geometric evidence that the axes are available directions in representation space, not that SGD explicitly optimizes either one as a standalone objective.
This gradient view also clarifies why some classical pruning scores behave like local-axis scores even when they are motivated by loss sensitivity.
Taylor importance $|a_i \cdot \nabla a_i|$, for example, is a first-order proxy for loss change.
Under the Gaussian model the task-loss gradient points along $\mathrm{cov}(X, T)$, but the \emph{magnitude} of that gradient at a given channel scales with $\|w_i\|$ and the projected input power $\RQ_i$, so the resulting Taylor score usually correlates more strongly with $I_X$ than with $I(T;Y)$ (per-architecture pooled $\rho(\text{Taylor}, I_X) \approx 0.48/0.39/0.30$ for ResNet-18/VGG-16/MobileNetV2, with a weaker but nonzero target component on MobileNetV2; PID-regression $R^2 < 0.22$).
This reinterprets Molchanov-style pruning~\citep{molchanov2019importance} as partly local-axis-sensitive, and helps explain why more direct local-axis scores can outperform it under the fixed pruning protocol in~\S\ref{sec:pruning}.

\subsection{Axis-specific information weakly propagates across layers}

After establishing that the axes separate within layers, we next ask whether this organization is purely local or whether it leaves a detectable trace across adjacent layers.
The result is intermediate.
Continuous axis values propagate weakly but consistently through weight connectivity: local$\to$local correlations fall in $\bar\rho=0.12$--$0.23$, target$\to$target correlations in $\bar\rho=0.09$--$0.21$, and cross-axis terms stay near $\bar\rho\approx0$, indicating that the two axes carry weakly conserved coordinates rather than fixed channel identities.
This matters because it shows that the two axes are not isolated layerwise artifacts: weak axis-specific coordinates are carried by weights, while identity-like channel types are rebuilt locally (Figure~\ref{fig:learned_main}E--H; full taxonomy summaries in Appendix~\ref{app:crosslayer_extra}).

The same local-axis view explains why magnitude is a strong baseline: $I_X$ contains a norm-weighted projected-power term.
At the same time, norm only moderately tracks $I_X$ and is weakly aligned with peer overlap (Appendix Table~\ref{tab:norm_local_boundary}), so local replaceability does not reduce to magnitude.
Matched-pair tests and a uniform-allocation sweep (Appendix~\ref{app:metric_weight_sweep}) show that local information adds value beyond norm, although the best norm-local combination is architecture-dependent.

\begin{figure}[!t]
\centering
\includegraphics[width=0.98\textwidth]{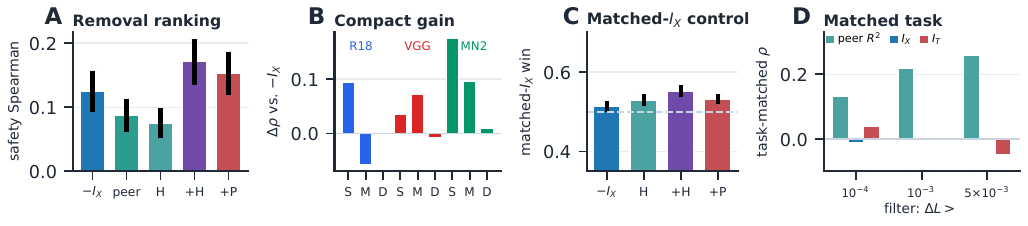}
\caption{\textbf{Direct lesion evidence.}
Larger scores predict lower lesion damage. Score key: $-I_X$ flips the input-capture proxy so low capture is high score; \emph{peer} is the within-layer overlap $\bar R_X$; $H$ is the compact-hull support score $E_i^\mathrm{full}/\max(1,|H_i|)$ (Appendix~\ref{app:computation_map} defines $E_i^\mathrm{full}$ and the greedy hull $H_i$); $+H$ is the within-cell standardized sum $z(-I_X)+z(H)$; $+P$ adds the top-8 peer-reconstruction $R^2$ in place of $H$. Depth labels S/M/D are shallow/mid/deep relative to total network depth.
Compact support improves ranking (A,B) and matched-$I_X$ win rate (C).
Panel D shows residual Spearman correlations with peer-replacement recovery within task-MI bins after positive-damage filtering; peer-reconstruction $R^2$ is the only consistently positive predictor among the tested scalar quantities.
R18 and MN2 denote ResNet-18 and MobileNetV2.
Full probes are in Appendix~\ref{app:hulls}.}
\label{fig:lesion_compact}
\end{figure}

\section{Intervention Tests via Pruning}
\label{sec:pruning}

If the local axis captures replaceability, it should predict what happens when channels are removed.
We therefore treat pruning as an intervention: does removal damage follow task relevance, or the availability of same-layer peers that can compensate?

\subsection{Single-channel lesions make replaceability visible}
\label{sec:lesion_replaceability}

We begin with the most direct intervention: removing individual channels without fine-tuning.
Across 45 model/seed/layer cells (2720 sampled channels total), we ablate sampled channels and measure loss increase on the fixed CIFAR-100 evaluation split.
For each channel, a compact peer explanation is the smallest same-layer peer set whose conditional Gaussian reconstruction nearly matches the full peer set.

Raw hull size is too crude: large hulls can mark broad coalitions rather than easy-to-remove channels, and its matched-$I_X$ win rate stays near chance (0.474; dense probe in Appendix~\ref{app:hulls}).
Compact peer support is the useful refinement (Figure~\ref{fig:lesion_compact}A--C), improving agreement with lower lesion damage ($0.171 \pm 0.036$ vs.\ $0.125 \pm 0.032$, mean $\pm$ SEM) and matched-$I_X$ win rate ($0.552 \pm 0.015$ vs.\ $0.513 \pm 0.014$; dense probe in Appendix~\ref{app:hulls}; task-matched Wilson CIs in Table~\ref{tab:matched_task_lesion_control}).
The task-matched control makes the central intervention claim explicit (Figure~\ref{fig:lesion_compact}D): within matched task-MI bins, peer-reconstruction $R^2$ predicts recovered lesion damage ($\rho=0.218$, 95\% CI $[0.152,0.282]$ for $\Delta L>10^{-3}$), while $I_X$ and task MI do not ($0.006$ and $-0.004$; Appendix Table~\ref{tab:matched_task_lesion_control}).
An explicit peer-replacement probe agrees: for nontrivial lesions ($\Delta L>10^{-3}$), top-8 peer reconstruction helps in 84.3\% of cases, recovers a median 68.7\% of the damage, and is predicted better by peer $R^2$ than by task MI ($\rho=0.423$ vs.\ $0.225$; Appendix Table~\ref{tab:peer_replacement_control}).
The effects are modest but consistent: relevance and input capture do not by themselves determine removability; peer replaceability changes lesion damage.

\subsection{Local replaceability is more predictive under matched pruning}
\label{sec:ixy_vs_taskmi}

We next move from lesions to structured pruning curves.
We group scores by the information they use: local-axis scores use $I_X$ and/or same-layer peer support $\bar R_X$, whereas target-axis scores use $I(T;Y)$, $\Red_T$, and/or $\Syn$.
All methods use the same global-threshold rule and architecture-aware FLOPs-matched accuracy-retention AUC over common model-specific FLOPs intervals; exact score definitions are in Appendix~\ref{app:experiments}.

\begin{figure}[!t]
    \centering
    \includegraphics[width=0.96\textwidth]{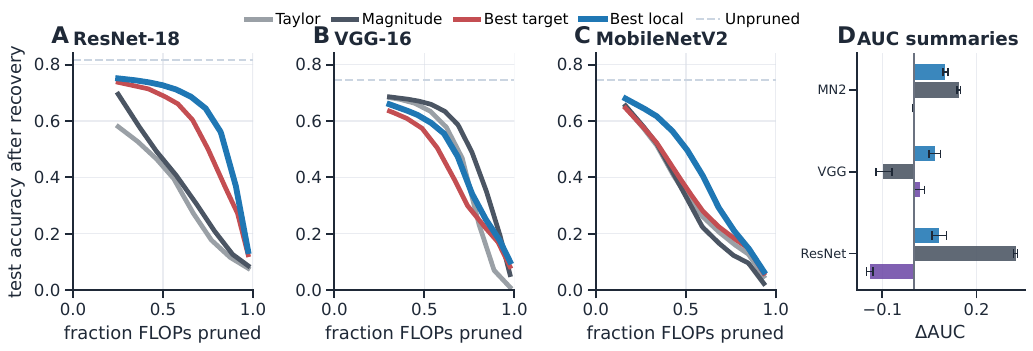}
    \caption{\textbf{FLOPs-matched pruning.}
    AUC uses common model-specific FLOPs intervals.
    Dashed lines in A--C show unpruned accuracy.
    In D: blue = local$-$target, orange = local$-$magnitude, purple = hybrid$-$local; horizontal bars show paired bootstrap 95\% CIs, MN2 denotes MobileNetV2, and magnitude remains strongest on VGG-16.}
    \label{fig:pruning_intervention}
\end{figure}

The best local-axis method exceeds the best target-axis method on ResNet-18 ($+8.0$pp; 95\% CI $[+5.7,+10.4]$), VGG-16 ($+6.6$pp $[+4.9,+8.6]$), and MobileNetV2 ($+10.1$pp $[+9.3,+11.0]$).
Against the strongest non-local baseline, best local remains ahead on ResNet-18 ($+32.4$pp $[+31.7,+33.1]$) and MobileNetV2 ($+14.4$pp $[+13.7,+14.9]$), but not VGG-16, where magnitude is stronger ($-9.7$pp $[-12.0,-6.9]$).
Breadth checks and random/FPGM/CHIP/Network-Slimming controls preserve the local-over-target hierarchy (Appendices~\ref{app:evidence} and~\ref{app:benchmark_pruning_extension}).
The modern-CNN stress test points in the same intervention direction: on ConvNeXt-T/ImageNet-100, best local exceeds best target in all 3 seeds ($+0.107 \pm 0.005$ AUC), although H1 is more coupled ($\rho=0.154 \pm 0.011$), so we treat it as supportive rather than primary evidence.
\section{Discussion}
\label{sec:discussion}

This work separates two properties often collapsed into channel importance: task relevance, or what a channel says about the target, and local replaceability, or whether same-layer peers can support its function when it is removed.
Across structural analyses, training trajectories, lesions, and FLOPs-matched pruning, trained networks weakly align the two axes, learning drives their separation, and removal damage follows local support more reliably than target relevance alone.
The conclusion is axis-level rather than leaderboard-level: under the fixed pruning protocol, local replaceability is a better guide to removability than target relevance, while the strongest scalar score remains architecture-dependent.
VGG-16 is the clearest boundary case: its Rayleigh-quotient variation is much smaller than MobileNetV2's (Appendix Table~\ref{tab:norm_local_boundary}), so weight norm can be a useful local-axis surrogate even though it misses peer overlap.
The framework therefore complements pruning heuristics, feature selection, and information-bottleneck views by asking a missing local question: among channels with similar target evidence, which are locally distinctive, and which are supported by peers?
The same distinction may transfer to sparse-autoencoder analyses of transformer representations~\citep{elhage2022toy,bricken2023towards,cunningham2024sparse,templeton2024scaling}, where dictionary features could be scored by target relevance and local replaceability; this remains a hypothesis, not a result of the present paper.

\paragraph{Limitations.}\label{par:limitations}
The strongest evidence comes from CIFAR-100 CNNs; CIFAR-10, Tiny-ImageNet, ImageNet-100, ConvNeXt, and ViT analyses are stress tests rather than complete architecture-general validation.
The Gaussian-MI quantities are structural proxies, not literal finite mutual-information estimates for deterministic continuous channels, and Proposition~\ref{prop:grad_orth} is limited to a linear-Gaussian mechanism with a stop-gradient target.
Pruning rankings also depend on recovery budget, score normalization, and cross-layer allocation, and local replaceability is measured primarily within layers rather than through all downstream compensatory paths.
The paper therefore supports a relevance--replaceability distinction rather than a universal pruning method.

\paragraph{Broader impact.}\label{par:broader_impact}
Better channel diagnostics may reduce inference cost, which can broaden access to efficient models but also lower the cost of harmful or low-accountability deployment.
This paper introduces no deployed model, private data, or user-facing capability; its contribution is diagnostic: task relevance explains what a channel contributes, while local replaceability explains whether the network still needs it.


\clearpage
\bibliographystyle{plainnat}
\IfFileExists{references.bib}{\bibliography{references}}{%
\bibliography{drafts/alignment_notes/references}}

\appendix
\AppendixFigurePacing
\paragraph{Appendix guide.}
The appendix is organized thematically as a support package for the four main claims.
H1 denotes weak alignment between the local and target axes; H2 denotes learned separation and the Gaussian residualization mechanism; H3 denotes weak axis-specific cross-layer propagation; and H4 denotes the intervention result that local-axis metrics predict removability better than target-axis metrics under the fixed protocol.
The thematic organization is:
\textbf{Notation and theory} (Appendix~\ref{app:computation_map}, Appendix~\ref{app:math}) defines the scalar and pairwise quantities used in every figure and gives the Gaussian-MMI collapse and residualized-gradient calculations.
\textbf{Protocol} (Appendix~\ref{app:experiments}) gives the fixed training, metric, and pruning conventions.
\textbf{Primary controls} (Appendices~\ref{app:practical}, \ref{app:controls}, \ref{app:evidence}) cover calibration sensitivity, target-definition and BatchNorm reparameterization controls, the full pruning tables with paired uncertainty, and the H1--H4 evidence matrix.
\textbf{Breadth and scope} (Appendices~\ref{app:benchmark_consistency}--\ref{app:convnext_pilot}, \ref{app:imagenet_recovery}) report cross-dataset structural consistency, the 11-cell pruning breadth suite, the ConvNeXt-T/ImageNet-100 stress test, and the ResNet-50/ImageNet-100 recovery-budget sensitivity check.
\textbf{Mechanism details} (Appendices~\ref{app:metric_weight_sweep}, \ref{app:pairwise}, \ref{app:crosslayer_extra}, \ref{app:theory_figure}) cover the uniform-allocation weight sweep, the pairwise/hull/triplet structure underlying Figure~\ref{fig:higher_order_depth} and the lesion analysis, the cross-layer propagation definition behind Figure~\ref{fig:learned_main}E--H, and the BROJA target-side decomposition.
\textbf{Pilots} (Appendix~\ref{app:teaser} prior 1-D vs.\ two-axis example and Appendix~\ref{app:rule_experiments} decorrelation regularizer and ViT pilot) mark exploratory extensions rather than primary evidence.
If reading selectively: Appendix~\ref{app:computation_map} for notation, Appendix~\ref{app:math} for theory, Appendix~\ref{app:experiments} for protocol, and Appendices~\ref{app:controls}--\ref{app:evidence} for the primary controls and the claim-to-evidence matrix.

\section{Notation and Figure-Computation Map}
\label{app:computation_map}

This appendix collects the quantities used across the main and supplementary figures in one place.
All scores are computed post hoc on frozen networks unless a training-trajectory section explicitly says otherwise.
For each convolutional layer, $Y_i$ denotes the Conv2d forward-hook output for channel $i$ (post-convolution, pre-BatchNorm, pre-ReLU, and pre-residual addition in the main convention), and target-side metrics use global-average-pooled channel activations.
Local pairwise correlations treat image and spatial positions as samples unless a section explicitly uses pooled activations.

\paragraph{Scalar two-axis quantities.}
The local scalar input-capture proxy is $I_X(i)=\frac{1}{2}\log(1+s_i/\sigma_0^2)$ with $s_i=w_i^\top\Sigma_Xw_i$ and per-layer $\sigma_0^2=\mathrm{median}_i(s_i)$.
The local peer-overlap coordinate is $\bar R_X(i)=(N_\ell-1)^{-1}\sum_{j\neq i}g(\rho_{ij})$, where $g(\rho)=-\frac{1}{2}\log(1-\rho^2)$.
The target singleton coordinate is $I(T;Y_i)=g(\rho_{T,i})$, where $T=f_z(x)-\max_{c\neq z}f_c(x)$ is the ground-truth logit margin in the main CNN experiments.
The scalable target-redundancy coordinate is $\Red_T(i)=m_i^{-1}\sum_{j\in\mathrm{Top}\text{-}m(i)}\min\{I(T;Y_i),I(T;Y_j)\}$.
The target-excess coordinate is $\Syn_i=m_i^{-1}\sum_{j\in\mathrm{Top}\text{-}m(i)}[I(T;[Y_i,Y_j])-\max\{I(T;Y_i),I(T;Y_j)\}]$, with $m_i=\min(10,N_\ell-1)$ by default.
Appendix Table~\ref{tab:synergy_partner_control} varies both $m$ and the partner-selection rule.

\paragraph{Pairwise and higher-order objects.}
For each layer, the local redundancy graph $R$ has edge weight
$R_{ij}=g(\rho_{ij})$.
The target-excess graph $S$ has edge weight
$S_{ij}=I(T;[Y_i,Y_j])-\max\{I(T;Y_i),I(T;Y_j)\}$, using the same Gaussian regression $R^2$ formula as Eq.~\ref{eq:syn}.
When reporting graph modularity, we retain the top 10\% of positive within-layer edges and compute greedy Newman modularity~\citep{newman2006modularity}; this is the source of the R-graph/S-graph comparisons in Figure~\ref{fig:higher_order_depth}A and Appendix~\ref{app:higher_order_training}.

For replaceability hulls, the peer explanation of channel $i$ by a peer set $S$ is the Gaussian linear-regression explained variance
\[
    E_i(S)=\rho_{iS}^{\top}\bigl(\Sigma_{SS}+10^{-6}I\bigr)^{-1}\rho_{iS},
\]
clipped to $[0,1]$.
The full peer explanation $E_i^\mathrm{full}$ is computed from all peers, or from the top peer pool in the dense lesion probe for tractability.
The greedy hull $H_i$ is the smallest selected peer set, capped at 10 channels, whose explanation reaches $(1-\varepsilon)E_i^\mathrm{full}$ with $\varepsilon=0.05$.
In the depth summaries, hull selection is restricted to the most correlated same-layer peer candidates; in the dense lesion probe, the same greedy rule is recomputed for the sampled channels.
The compact-hull score used in Figure~\ref{fig:lesion_compact} is $E_i^\mathrm{full}/\max(1,|H_i|)$, and the local compact score is the within-cell standardized sum $z(-I_X)+z(E_i^\mathrm{full}/\max(1,|H_i|))$.
Higher scores in the lesion figure are oriented as safer-to-remove scores.

Triplet target-excess is computed on the top 24 task-relevant channels per layer.
For a sampled triplet $(i,j,k)$,
\[
    S_3(T;i,j,k)=I(T;[Y_i,Y_j,Y_k])-\max_{(a,b)\subset(i,j,k)} I(T;[Y_a,Y_b]),
\]
and the plotted ratio $S_3/S_2$ divides the mean clipped triplet excess by the corresponding mean pair excess $S_2=\max_\mathrm{pair}I(T;[Y_a,Y_b])-\max\{I(T;Y_i),I(T;Y_j),I(T;Y_k)\}$.
All joint MI terms are Gaussian regression proxies based on the coefficient of determination of $T$ on the selected channels.

\paragraph{Figure map.}
Figure~\ref{fig:schematic} is the conceptual summary.
Figure~\ref{fig:two_axes} uses the scalar two-axis quantities above and the Gaussian-MMI target-redundancy collapse.
Figure~\ref{fig:higher_order_depth} uses R/S graph modularity, hull-size summaries, singleton/saturated hull fractions, and $S_3/S_2$.
Figure~\ref{fig:learned_main} recomputes the same scalar axes on checkpointed networks and uses the weight-routed cross-layer metric flow defined in Eq.~\ref{eq:weighted_metric}.
Figure~\ref{fig:lesion_compact} uses single-channel lesions, compact-hull support, and top-8 peer replacement controls defined in Appendix~\ref{app:hulls}.
Figure~\ref{fig:pruning_intervention} uses the local/target pruning score families and the FLOPs-matched AUC protocol defined in Appendix~\ref{app:experiments}.

\section{Mathematical Details}
\label{app:math}

\subsection{Gaussian MI formulas}

For jointly Gaussian $(Y_i, Y_j)$ with correlation $\rho$:
\begin{equation}
    I(Y_i; Y_j) = -\tfrac{1}{2}\log(1 - \rho^2).
\end{equation}

For jointly Gaussian $(T, Y_i)$ with correlation $\rho_{Ti}$:
\begin{equation}
    I(T; Y_i) = -\tfrac{1}{2}\log(1 - \rho_{Ti}^2).
\end{equation}

For jointly Gaussian $(T, Y_i, Y_j)$:
\begin{equation}
    I(T; [Y_i, Y_j]) = -\tfrac{1}{2}\log(1 - R^2),
\end{equation}
where $R^2$ is the coefficient of determination from regressing $T$ on $(Y_i, Y_j)$.

\subsection{MMI PID collapse proof}

The MMI redundancy is:
$\Red_\text{MMI}(T; Y_1, Y_2) = \min\{I(T;Y_1), I(T;Y_2)\}$.

The PID identity gives:
\begin{align}
    I(T; [Y_1, Y_2]) &= \text{Uniq}_1 + \text{Uniq}_2 + \Red + \Syn,\\
    I(T; Y_1) &= \text{Uniq}_1 + \Red,\\
    I(T; Y_2) &= \text{Uniq}_2 + \Red.
\end{align}

Without loss of generality, let $I(T;Y_1) \leq I(T;Y_2)$.
Then $\Red = I(T;Y_1)$, so $\text{Uniq}_1 = 0$ and $\text{Uniq}_2 = I(T;Y_2) - I(T;Y_1)$.
Thus, $\Red = I(T;Y_1)$ under this ordering and $\text{Uniq}_2$ is also determined by individual task MI values.

This means that, under MMI, target-side redundancy and unique-information terms are determined by singleton task MI values.
Pairwise synergy still depends on the joint information $I(T;[Y_1,Y_2])$ and is treated separately in the main text.
Empirically, the resulting target-redundancy scores correlate at $\rho \geq 0.994$ with singleton task MI across the canonical CIFAR-100 layers.

\subsection{Gradient residualization (derivation for Prop.~\ref{prop:grad_orth})}
\label{app:math_grad}

Consider the Gaussian channel $Y_i = w_i^\top X + \varepsilon$ with $X \sim \mathcal{N}(0, \Sigma_X)$ and $\varepsilon \sim \mathcal{N}(0, \sigma_0^2)$ independent of $X$.
Let $s_i = w_i^\top \Sigma_X w_i$ denote the signal power.
The input-capture proxy is
\[
I_X(i) \;=\; \tfrac{1}{2} \log\!\bigl(1 + s_i/\sigma_0^2\bigr),
\]
so
\[
\frac{\partial I_X(i)}{\partial w_i}
\;=\;
\frac{1}{2}\frac{1}{1 + s_i/\sigma_0^2}\cdot\frac{\partial (s_i/\sigma_0^2)}{\partial w_i}
\;=\;
\frac{\Sigma_X w_i}{\sigma_0^2 + s_i}.
\]
For the target-axis proxy let $T$ be a zero-mean task variable with $\rho_{T,i} = \mathrm{corr}(T, Y_i)$, treated as external to the local channel derivative.
Writing $b_i=w_i^\top c$, $D_i=s_i+\sigma_0^2$, $\rho_{T,i}^2 = b_i^2 / (D_i \cdot \sigma_T^2)$ with $c = \mathrm{cov}(X, T)$ and $\sigma_T^2 = \Var(T)$,
\[
I(T; Y_i) = -\tfrac{1}{2}\log(1 - \rho_{T,i}^2),
\qquad
\frac{\partial I(T;Y_i)}{\partial w_i}
\;=\;
\frac{1}{1 - \rho_{T,i}^2}\;\frac{b_i}{D_i\,\sigma_T^2}\;
\Bigl(\,c \;-\; \frac{b_i}{D_i}\, \Sigma_X w_i\Bigr),
\]
so
\[
\frac{\partial I(T;Y_i)}{\partial w_i}
\;\propto\;
c - \frac{b_i}{D_i}\,\Sigma_X w_i
\]
up to the scalar $(1 - \rho_{T,i}^2)^{-1}\,b_i/(D_i\,\sigma_T^2)$.
The first gradient points along $\Sigma_X w_i$ (the signal-power direction), the second along $c - (b_i/D_i) \Sigma_X w_i$ (a residualized task-covariance direction).
Their (unnormalised) inner product is
\[
\Bigl\langle \Sigma_X w_i,\; c - \frac{b_i}{D_i}\,\Sigma_X w_i\Bigr\rangle
\;=\; w_i^\top \Sigma_X c \;-\; \frac{b_i}{D_i}\,w_i^\top \Sigma_X^2 w_i,
\]
which vanishes whenever the projection of $c$ onto $w_i$ through $\Sigma_X$ cancels the squared signal-power term.
Generically, when the task-covariance direction $c$ is weakly aligned with the dominant input-power directions of $\Sigma_X$, the two gradients can have small expected cosine across channels.
At the final checkpoint, the measured update-direction cosine is small but positive, $\cos(-\nabla L,\, \nabla I_X)=0.061\pm0.009$ (Section~\ref{sec:learned}); equivalently, the loss gradient has a small negative local-axis component.
This should be read as a sign-sensitive geometric diagnostic, not as evidence that SGD explicitly optimizes $I_X$ as a standalone objective.
\qed

\section{Experimental Details and Fixed Protocol}
\label{app:experiments}

\paragraph{Training.}
All models are trained with SGD (momentum 0.9, weight decay $5 \times 10^{-4}$) for 200 epochs with cosine annealing.
Initial learning rate: 0.1 (ResNet, VGG) or 0.05 (MobileNet).
Standard CIFAR augmentation: random crop (32$\times$32, padding 4), horizontal flip.
Unless a section explicitly marks a pilot or sweep with fewer checkpoints, canonical CIFAR-100 results use seeds 42, 123, 456, 789, and 1011.

\paragraph{Datasets, software, and assets.}
The main experiments use CIFAR-100, with additional CIFAR-10 breadth checks~\citep{krizhevsky2009learning}.
Tiny-ImageNet~\citep{tinyimagenet} is used as a small ImageNet-derived stress test, and ImageNet-100 subsets inherit the ImageNet access and usage terms~\citep{deng2009imagenet}.
Implementations use PyTorch and torchvision~\citep{paszke2019pytorch,torchvision2016}; the ConvNeXt-T stress test uses the ConvNeXt architecture~\citep{liu2022convnext}.
No new dataset or pretrained model is introduced.
The experiments were run through the project codebase, which provides model wrappers, activation hooks, metric extraction, pruning operators, FLOPs accounting, and checkpoint orchestration.
The companion code release documents command-level reproduction; the appendix here records the scientific protocol and analysis definitions.

\paragraph{Metric computation.}
Calibration uses a fixed 5000-image subset of the training split, stored as explicit indices and reused across seeds and metrics.
Convolutional inputs are unfolded into patches before estimating $\Sigma_X$ for $I_X$; output-side local correlations use the stored Conv2d forward-hook activations (post-convolution, pre-BN, pre-ReLU, and pre-residual-addition in the main runs), treating image and spatial positions as calibration samples.
Target-side correlations use global-average-pooled channel activations so that each image contributes one scalar per channel.
For $I_X$, $\RQ_i=w_i^\top\Sigma_Xw_i/\|w_i\|^2$ and $\sigma_0^2=\mathrm{median}_i(\RQ_i\|w_i\|^2)$ are computed separately within each layer in the main runs.
$\bar{R}_X$ uses the full same-layer channel-pair correlation matrix, not a sampled or top-$K$ neighborhood, converts each pair to Gaussian MI, removes the diagonal, and averages over all peers.
$I(T;Y)$ correlates each GAP-pooled channel with the ground-truth logit margin $T=f_z(x)-\max_{c\neq z} f_c(x)$.
$\Syn$ uses top-10 singleton-task-MI partners per channel as the default pairwise Gaussian-MMI target-excess proxy; Appendix~\ref{app:controls} reports partner-count and partner-selection controls.
All Pearson correlations entering Gaussian MI formulas are clipped away from $\pm1$ for numerical stability, and zero-variance channels are excluded from rank/correlation summaries for the affected layer.

\paragraph{Pruning protocol.}
For each canonical CIFAR-100 pruning method and the additional controls, we evaluate 10 sparsity levels (10\%--95\% of channels removed).
After pruning, we fine-tune with SGD (lr=0.01, momentum 0.9, weight decay $10^{-4}$) for 5 epochs, capped at 200 optimizer batches under the fixed pruning protocol used by the canonical suite and fixed-protocol breadth extension.
Appendix~\ref{app:imagenet_recovery} reports a separate historical ResNet-50/ImageNet-100 recovery-budget sensitivity check comparing a 50-batch run with a 200-batch rerun.
FLOPs are computed architecture-aware via torch.fx symbolic tracing, accounting for output-channel removal and cascading input-channel reduction through depthwise convolutions and residual connections.
AUC is computed over a common FLOPs range per model where all methods have data.
All results are averaged over 5 random seeds (42, 123, 456, 789, 1011).

\paragraph{Baselines and methods.}
Taylor (first-order)~\citep{molchanov2019importance}, magnitude ($\ell_2$ weight norm), random channel pruning~\citep{li2022random}, geometric-median/FPGM redundancy~\citep{he2019filter}, CHIP channel independence~\citep{sui2021chip}, and mutual-information preserving pruning (MIPP) as the closest adjacent-layer MI prior~\citep{westphal2024mipp}.
Composite-$I_X$ uses $\alpha=2.0$, $\gamma=0.25$, and no task-MI component.
CAP-QGW denotes the cluster-aware pruning variant that combines local-axis rankings with quantile and gradient-weighted allocation; CAP-A denotes the annealed cluster-aware variant used in the ImageNet-100 recovery-budget check.
PID-based rankings, spectral clustering variants using pairwise redundancy (R) or synergy (S) affinity.
Hybrid: Taylor scores for global threshold (cross-layer allocation), $I_X$ for within-layer ranking.
The fixed comparisons are interpreted at the axis level; the spectral and synergy-based variants are included for completeness rather than as main methods.
Where a baseline is not part of the main five-method figure, we report it in the breadth or appendix tables and keep the claim at the family level rather than as a SOTA pruning leaderboard.

\section{Practical and Robustness Checks}
\label{app:practical}

\paragraph{Calibration-set sensitivity.}
The main results use a fixed 5000-image calibration subset to keep all metrics on one fixed protocol.
To assess sensitivity to calibration size, we recomputed the two-axis scalar summary $(I_X, \bar R_X, I(T;Y), \Syn)$ on smaller subsets and compared the resulting layerwise rankings against the 5000-image reference.
Rank stability rises with calibration size and saturates well before the 5000-image budget; the qualitative two-axis separation is already visible at smaller sizes (Table~\ref{tab:calibration_sensitivity}).
We use 5000 images in the main runs because it remains a single forward-pass-only post-hoc computation while keeping calibration noise below the cross-axis effect size.

\begin{table}[htbp]
\centering
\caption{\textbf{Calibration-size sensitivity of the two-axis scalar summary.}}
\label{tab:calibration_sensitivity}
\small
\begin{tabular}{lcc}
\toprule
$n_{\mathrm{cal}}$ & ARI vs full & Mean silhouette \\
\midrule
100 & 0.295 $\pm$ 0.006 & 0.298 $\pm$ 0.002 \\
500 & 0.322 $\pm$ 0.009 & 0.301 $\pm$ 0.002 \\
1000 & 0.344 $\pm$ 0.010 & 0.303 $\pm$ 0.002 \\
2000 & 0.363 $\pm$ 0.013 & 0.303 $\pm$ 0.005 \\
5000 & 1.000 $\pm$ 0.000 & 0.310 $\pm$ 0.000 \\
\bottomrule

\end{tabular}
\end{table}

\paragraph{Practical notes.}
Metric extraction is a post-hoc computation rather than a retraining loop: one frozen-model forward pass over the fixed calibration set, followed by within-layer covariance and correlation statistics.
It does not require repeated optimization, recovery tuning, or multiple calibration resweeps.
To make this concrete, Table~\ref{tab:runtime_cost} summarizes measured wall-clock ranges from the 53 breadth-suite jobs on one H100-class GPU.
The ``metric extraction'' column is measured directly from the run logs as the interval between ``Computing per-channel metrics'' and ``Computing per-channel loss proxy,'' so it isolates the post-hoc structural computation rather than the later pruning sweep.
Across the broader benchmark layer, metric extraction stays in the 0.8--3.2 minute range for the CIFAR-10 and Tiny-ImageNet CNNs, and in the 1.1--12.7 minute range for the ImageNet-100 family.
The full fixed-protocol reruns are much more expensive because they include pruning, capped recovery, and evaluation, ranging from 16.9 to 106.1 minutes depending on dataset and backbone.
Thus, the dominant incremental cost of using the metrics as analysis tools is the pruning intervention, not the metric extraction itself.

\begin{table}[htbp]
\centering
\caption{\textbf{Measured wall-clock cost from the fixed breadth suite (one H100-class GPU).}}
\label{tab:runtime_cost}
\small
\begin{tabular}{lcc}
\toprule
Benchmark family & Metric extraction & Full fixed run \\
\midrule
CIFAR-10 CNNs & 0.8--1.3 min & 16.9--26.5 min \\
Tiny-ImageNet CNNs & 1.8--3.2 min & 24.9--37.2 min \\
ImageNet-100 family & 1.1--12.7 min & 37.2--106.1 min \\
\bottomrule
\end{tabular}
\end{table}

\section{Control Experiments}
\label{app:controls}

These controls are the direct structural checks for H1--H3 and the metric-estimation part of H4.
The pruning controls C5--C9 are reported in Appendix~\ref{app:evidence}; the target-definition and BatchNorm reparameterization checks below address common concerns about target choice and parameterization.
Later appendix sections provide interpretation and scope checks, but are not additional main claims.

\paragraph{C1: Weak cross-axis alignment.}
The figure-level layer summary gives within-layer Spearman $\rho(I_X, I(T;Y)) = 0.012 \pm 0.08$ (mean $\pm$ SD over 425 layer$\times$seed pairs) and ARI between local and target clusterings $0.020 \pm 0.04$, versus null mean $0.001 \pm 0.02$.
Model-seed bootstrap and leave-one-architecture-out aggregation preserve the weak-alignment conclusion while shifting the exact point estimate because the weighting unit changes (Table~\ref{tab:hierarchical_robustness}).
PASS.

\begin{table}[htbp]
\centering
\caption{\textbf{Hierarchical robustness for weak alignment.}
Bootstrap intervals resample model-seed units rather than treating individual layers as independent; leave-one-architecture-out means remain near the same weak-alignment regime.}
\label{tab:hierarchical_robustness}
\small
\resizebox{\linewidth}{!}{%
\begin{tabular}{lccccc}
\toprule
Statistic & Bootstrap mean [95\% CI] & Layer-cell mean $\pm$ SEM & w/o R18 & w/o VGG & w/o MN2 \\
\midrule
$\rho(I_X,I(T;Y))$ & $-0.016~[-0.028,-0.002]$ & $-0.016 \pm 0.008$ & $-0.021$ & $-0.012$ & $-0.012$ \\
ARI(local,target) & $0.035~[0.029,0.042]$ & $0.031 \pm 0.003$ & $0.026$ & $0.031$ & $0.041$ \\
\bottomrule
\end{tabular}
}
\end{table}

\paragraph{C2: PID collapse.}
Pooled $\rho(I(T;Y), \Red_T)=0.990$ in Figure~\ref{fig:two_axes}B; per-layer summaries satisfy $\rho(I(T;Y), \Red_T) \geq 0.994$ across the canonical models.
$\rho(\bar{R}_X, \Red_T) \approx 0.05$.
PASS.

\paragraph{C2b: Empirical target-side robustness.}
Auxiliary nonparametric target-side checks preserve the same broad pattern.
Mean model-level correlation between empirical BROJA-style redundancy summaries and task MI is 0.883 (ResNet-18), 0.884 (VGG-16), and 0.812 (MobileNetV2), while Gaussian vs.\ KSG channelwise task-MI ranks are weakly aligned (mean $\rho = 0.099$, 0.015, 0.103 respectively).
This supports the collapse as a structural relation among target-side quantities rather than a claim of channelwise estimator equivalence.
PASS with caveat.

\paragraph{C3: Label permutation.}
Under random label permutation: task MI drops $50\times$; $I_X$ unchanged.
Confirms that task metrics are target-specific while local metrics are target-independent.
PASS.

\paragraph{C4: $I_X$ estimator robustness.}
Channel rankings by $I_X$ are preserved ($\tau > 0.95$) between exact covariance and streaming RQ estimation modes.
PASS.

\paragraph{Target-definition sensitivity.}
The target axis uses the ground-truth logit margin in the main analyses, but the weak local-target alignment is not an artifact of that target variable.
Across the same 3 backbones $\times$ 5 seeds, replacing the target by the correct-class logit, negative loss, predicted-class margin, predicted-class logit, or a one-vs-rest label target keeps $\rho(I_X,\mathrm{target})$ small (Table~\ref{tab:target_sensitivity_control}).
The predicted-class margin is almost identical to the ground-truth margin after training, whereas logit and loss targets define different channel rankings but still do not collapse onto the local axis.
Table~\ref{tab:target_agreement_matrix} shows the pairwise agreement among the target-information rankings themselves: the two margin definitions and the two logit definitions form near-duplicate pairs, but the margin, logit, loss, and label targets otherwise control different target-side notions.

\begin{table}[htbp]
\centering
\caption{\textbf{Target-definition sensitivity} (3 backbones $\times$ 5 seeds; mean $\pm$ SEM over model-seed summaries).
The local-target association remains weak across target choices, so the two-axis result is not specific to the ground-truth margin target.}
\label{tab:target_sensitivity_control}
\small
\begin{tabular}{lccc}
\toprule
Target variable & $\rho(I_X,\mathrm{target})$ & ARI local/target & $\tau$ vs.\ GT margin \\
\midrule
GT margin & $0.023 \pm 0.007$ & $0.013 \pm 0.002$ & $1.000 \pm 0.000$ \\
Correct-class logit & $0.043 \pm 0.014$ & $0.015 \pm 0.002$ & $0.230 \pm 0.056$ \\
Negative loss & $0.019 \pm 0.007$ & $0.010 \pm 0.001$ & $0.137 \pm 0.023$ \\
Predicted-class margin & $0.023 \pm 0.007$ & $0.012 \pm 0.002$ & $0.990 \pm 0.003$ \\
Predicted-class logit & $0.043 \pm 0.014$ & $0.015 \pm 0.003$ & $0.229 \pm 0.056$ \\
One-vs-rest label & $0.095 \pm 0.015$ & -- & $0.053 \pm 0.007$ \\
\bottomrule
\end{tabular}
\end{table}

\begin{table}[htbp]
\centering
\caption{\textbf{Agreement among target-information definitions} (pairwise Kendall $\tau$ between channel rankings; mean over 3 backbones $\times$ 5 seeds).
The target choices are not all interchangeable: margins and logits form two near-duplicate pairs, while loss and one-vs-rest label information induce different target-side rankings. Table~\ref{tab:target_sensitivity_control} shows that even these different target rankings remain weakly aligned with the local axis.}
\label{tab:target_agreement_matrix}
\scriptsize
\begin{tabular}{lrrrrrr}
\toprule
Target & GT marg. & Corr. logit & Neg. loss & Pred. marg. & Pred. logit & Label OVR \\
\midrule
GT margin & 1.000 & 0.230 & 0.137 & 0.990 & 0.229 & 0.053 \\
Correct logit & 0.230 & 1.000 & 0.039 & 0.231 & 0.994 & 0.032 \\
Negative loss & 0.137 & 0.039 & 1.000 & 0.134 & 0.038 & 0.003 \\
Predicted margin & 0.990 & 0.231 & 0.134 & 1.000 & 0.230 & 0.053 \\
Predicted logit & 0.229 & 0.994 & 0.038 & 0.230 & 1.000 & 0.032 \\
Label OVR & 0.053 & 0.032 & 0.003 & 0.053 & 0.032 & 1.000 \\
\bottomrule
\end{tabular}
\end{table}

\begin{table}[htbp]
\centering
\caption{\textbf{Target-synergy partner-selection robustness} (3 backbones $\times$ 5 seeds; mean $\pm$ SEM over model-seed summaries).
Changing the number of partners or selecting partners by joint MI, direct target excess, or random task-relevant peers preserves the same qualitative structure: the target-excess proxy remains target-side and weakly aligned with the local input-capture axis.}
\label{tab:synergy_partner_control}
\small
\resizebox{\linewidth}{!}{%
\begin{tabular}{lcccc}
\toprule
Partner rule & $\rho(I_X,\Syn)$ & $\rho(I(T;Y),\Syn)$ & $\tau$ vs.\ default & ARI local/target \\
\midrule
Top task MI, $m=5$ & $0.035 \pm 0.007$ & $0.634 \pm 0.017$ & $0.760 \pm 0.007$ & $0.013 \pm 0.002$ \\
Top task MI, $m=10$ & $0.043 \pm 0.008$ & $0.696 \pm 0.015$ & $1.000 \pm 0.000$ & $0.013 \pm 0.002$ \\
Top task MI, $m=20$ & $0.044 \pm 0.009$ & $0.729 \pm 0.013$ & $0.812 \pm 0.006$ & $0.012 \pm 0.002$ \\
Top joint MI, $m=10$ & $0.059 \pm 0.010$ & $0.730 \pm 0.011$ & $0.786 \pm 0.007$ & $0.018 \pm 0.002$ \\
Top synergy, $m=10$ & $0.059 \pm 0.013$ & $0.723 \pm 0.012$ & $0.707 \pm 0.009$ & $0.018 \pm 0.002$ \\
Random top-task pool, $m=10$ & $0.039 \pm 0.009$ & $0.747 \pm 0.011$ & $0.703 \pm 0.009$ & $0.012 \pm 0.002$ \\
\bottomrule
\end{tabular}
}
\end{table}

\paragraph{BatchNorm reparameterization control.}
For the BatchNorm backbones, raw convolutional parameter quantities are not invariant to function-preserving Conv--BN rescalings.
We therefore report $I_X$ as a fixed-representation diagnostic and compare pruning against BN-scale and activation-power baselines rather than treating them as equivalent forms of the same score.
In an artificial Conv--BN rescaling control, raw conv-norm rankings change strongly, while BN-folded norms and post-BN activation variances are preserved to Kendall $\tau\approx1$ despite small finite-precision logit differences (Table~\ref{tab:bn_reparam_control}).

\begin{table}[htbp]
\centering
\caption{\textbf{Conv--BatchNorm reparameterization control} (3 backbones $\times$ 5 seeds; mean $\pm$ SEM).
Artificial Conv--BN rescaling changes raw conv-space rankings while preserving BN-folded and post-BN quantities to near-perfect rank agreement; predictions are effectively unchanged under the finite-precision rescaling.}
\label{tab:bn_reparam_control}
\small
\begin{tabular}{lc}
\toprule
Quantity under artificial rescaling & Kendall $\tau$ or output change \\
\midrule
Maximum absolute logit difference & $0.0196 \pm 0.0049$ \\
Mean absolute logit difference & $0.00147 \pm 0.00060$ \\
Relative RMS logit difference & $0.00070 \pm 0.00021$ \\
Top-1 agreement & $0.9992 \pm 0.0004$ \\
Raw conv-norm ranking & $0.142 \pm 0.010$ \\
BN-folded norm ranking & $1.000 \pm 0.000$ \\
Raw conv-activation variance ranking & $0.260 \pm 0.004$ \\
Post-BN activation variance ranking & $1.000 \pm 0.000$ \\
\bottomrule
\end{tabular}
\end{table}

\section{Intervention Evidence and Full Pruning Tables}
\label{app:evidence}

\begin{table}[htbp]
\centering
\caption{\textbf{FLOPs-based accuracy retention AUC} (global-threshold pruning, 10 sparsity levels, AUC over common FLOPs range per model).
Methods are grouped by axis. Best non-hybrid method per model is in \textbf{bold}; the final row reports the best fixed-protocol hybrid and is bolded only if it exceeds the best non-hybrid score. Values are mean $\pm$ SEM over 5 seeds, except the descriptive hybrid row, which is reported as the fixed-suite mean.}
\label{tab:pruning_flops_appendix}
\small
\begin{tabular}{llccc}
\toprule
Axis & Method & ResNet-18 & VGG-16 & MobileNetV2 \\
\midrule
\multicolumn{5}{l}{\emph{Baselines}} \\
& Taylor & $0.402 \pm 0.008$ & $0.600 \pm 0.010$ & $0.442 \pm 0.003$ \\
& Magnitude & $0.454 \pm 0.005$ & $\mathbf{0.706 \pm 0.001}$ & $0.420 \pm 0.001$ \\
\midrule
\multicolumn{5}{l}{\emph{Local axis (replaceability)}} \\
& Composite-$I_X$ & $0.713 \pm 0.020$ & $0.592 \pm 0.004$ & $\mathbf{0.564 \pm 0.003}$ \\
& CAP-QGW & $\mathbf{0.774 \pm 0.005}$ & $0.611 \pm 0.013$ & $0.550 \pm 0.002$ \\
& Spectral-RS & $0.646 \pm 0.015$ & $0.561 \pm 0.007$ & $0.550 \pm 0.003$ \\
\midrule
\multicolumn{5}{l}{\emph{Target axis (task-directed PID)}} \\
& Composite-PID & $0.694 \pm 0.007$ & $0.504 \pm 0.009$ & $0.457 \pm 0.004$ \\
& Spectral-PID-RS & $0.675 \pm 0.009$ & $0.543 \pm 0.008$ & $0.464 \pm 0.005$ \\
\midrule
\multicolumn{5}{l}{\emph{Hybrid (allocation $\times$ ranking)}} \\
& Best hybrid (fixed) & $0.637$ & $0.632$ & $0.562$ \\
\bottomrule
\end{tabular}
\end{table}

Table~\ref{tab:pruning_delta_ci} reports paired bootstrap intervals for the main AUC gaps.

\begin{table}[htbp]
\centering
\caption{\textbf{Paired uncertainty for main pruning gaps.}
Entries are paired seed-level mean differences in FLOPs-AUC percentage points with 95\% bootstrap CIs.
The hybrid row is descriptive because the locked hybrid suite uses a different common FLOPs support from the main local/target suite.}
\label{tab:pruning_delta_ci}
\small
\resizebox{\linewidth}{!}{\begin{tabular}{lccc}
\toprule
Backbone & best local $-$ best target & best local $-$ magnitude & hybrid $-$ best local \\
\midrule
ResNet-18 & $+8.0$ [+5.7, +10.4] & $+32.4$ [+31.7, +33.1] & $-14.0$ [-15.1, -12.9] \\
VGG-16 & $+6.6$ [+4.9, +8.6] & $-9.7$ [-12.0, -6.9] & $+2.0$ [+0.1, +3.4] \\
MobileNetV2 & $+10.1$ [+9.3, +11.0] & $+14.4$ [+13.7, +14.9] & $-0.1$ [-0.3, +0.1] \\
\bottomrule
\end{tabular}
}
\end{table}

The raw pruning curves do not all begin at the same FLOPs fraction because the fixed channel-sparsity grid is converted to architecture-aware FLOPs after pruning.
For example, the first ResNet-18 point ranges from 0.024 FLOPs pruned for magnitude to 0.236 for Composite-PID, while the AUC comparison uses the common interval $[0.251,0.973]$.
Figure~\ref{fig:pruning_intervention} uses the same common intervals for visualization, avoiding apparent horizontal shifts from method-specific FLOPs support.

Table~\ref{tab:topup_controls} adds pruning controls under the same fixed checkpoint and recovery protocol.
Because the table includes additional methods, AUC is recomputed over the expanded common FLOPs-pruned range per model where all listed methods have support: ResNet-18 $[0.251,0.915]$, VGG-16 $[0.306,0.925]$, and MobileNetV2 $[0.165,0.933]$.
The absolute values therefore differ from Table~\ref{tab:pruning_flops_appendix}, but the control answers a different question: whether the local-axis result survives random, geometric-median, CHIP, BN-scale, and activation-power baselines under one common comparison window.

\begin{table}[htbp]
\centering
\caption{\textbf{Additional pruning controls} (FLOPs-based accuracy-retention AUC over the expanded common FLOPs range induced by all listed methods; mean $\pm$ SEM over 5 seeds).
The result supports the metric-family claim, not a universal SOTA pruning claim: best local-axis methods beat the best target-axis and CHIP methods on all three backbones, while VGG-16 remains a norm-like boundary where magnitude/FPGM/Network-Slimming-style baselines are very strong.}
\label{tab:topup_controls}
\small
\begin{tabular}{lccc}
\toprule
Method / family & ResNet-18 & VGG-16 & MobileNetV2 \\
\midrule
Best local-axis method & $\mathbf{0.669 \pm 0.004}$ & $0.483 \pm 0.012$ & $\mathbf{0.420 \pm 0.003}$ \\
Best target-axis method & $0.606 \pm 0.005$ & $0.448 \pm 0.008$ & $0.346 \pm 0.004$ \\
\midrule
Taylor & $0.349 \pm 0.008$ & $0.482 \pm 0.009$ & $0.329 \pm 0.002$ \\
Magnitude & $0.393 \pm 0.005$ & $\mathbf{0.563 \pm 0.001}$ & $0.313 \pm 0.001$ \\
Random channel pruning & $0.614 \pm 0.002$ & $0.523 \pm 0.003$ & $0.285 \pm 0.003$ \\
FPGM / geometric median & $0.396 \pm 0.005$ & $0.561 \pm 0.001$ & $0.315 \pm 0.002$ \\
CHIP & $0.130 \pm 0.002$ & $0.108 \pm 0.013$ & $0.284 \pm 0.002$ \\
Network Slimming & $0.546 \pm 0.005$ & $0.525 \pm 0.002$ & $0.267 \pm 0.002$ \\
Activation RMS & $0.243 \pm 0.003$ & $0.140 \pm 0.012$ & $0.320 \pm 0.001$ \\
\bottomrule

\end{tabular}
\end{table}

\paragraph{BatchNorm parameterization note.}
The main $I_X$ measurements are tied to the fixed convolutional-stage representation stored with each run.
They should not be read as a BatchNorm-invariant saliency score by themselves: the reparameterization control in Table~\ref{tab:bn_reparam_control} shows that raw conv-space rankings change under artificial Conv--BN rescaling, while BN-folded and post-BN quantities are stable.
For this reason we treat Network-Slimming-style BN scale and activation-RMS rankings as explicit baselines rather than as equivalent forms of $I_X$.
Table~\ref{tab:topup_controls} shows that the local-vs-target conclusion survives those controls, while also marking VGG-16 as the architecture where BN/norm-like saliency is strongest.

Table~\ref{tab:evidence} maps the main claims H1--H4 to the row-level controls C1--C9 that support them.

\begin{table}[htbp]
\centering
\caption{\textbf{Claim-to-evidence matrix.}
Main claim $\leftrightarrow$ row-level control, with data source, key metric, and pass/fail status.}
\label{tab:evidence}
\small
\begin{tabular}{@{}>{\raggedright\arraybackslash}p{0.11\linewidth}>{\raggedright\arraybackslash}p{0.20\linewidth}>{\raggedright\arraybackslash}p{0.17\linewidth}>{\raggedright\arraybackslash}p{0.33\linewidth}c@{}}
\toprule
Claim & Row-level control & Run family & Key metric & Status \\
\midrule
H1 & C1: weak cross-axis alignment & canonical 3-backbone suite & $\rho=0.012$, ARI$=0.020$ & PASS \\
H1 & Hierarchical aggregation & model-seed bootstrap & $\rho=-0.016$ [95\% CI: $-0.028,-0.002$]; ARI$=0.035$ & PASS \\
H1 & C2: PID collapse & canonical 3-backbone suite & pooled $\rho=0.990$; per-layer $\rho\geq0.994$ & PASS \\
H1 & C2b: Empirical target-side robustness & BROJA + KSG pipeline & BROJA vs.\ task: 0.81--0.88; Gaussian vs.\ KSG task-MI weak & PASS$^\dagger$ \\
H1 & Target-excess partner control & synergy robustness suite & $\rho(I_X,\Syn)\leq0.059$ across partner rules; ARI$\leq0.018$ & PASS \\
H1 & C3: Target permutation & target-permutation control & task MI drops $50\times$; $I_X$ unchanged & PASS \\
H1 & Target-definition sensitivity & target-control suite & $\rho(I_X,\mathrm{target}) \leq 0.095$ across six targets & PASS \\
H1/H2 & C4: $I_X$ estimator robustness & canonical 3-backbone suite & $\tau > 0.95$ & PASS \\
H1/H2 & BN reparameterization & parameterization-control suite & raw conv ranks change; BN-folded/post-BN $\tau\approx1$ & PASS \\
H2 & Gradient residualization & checkpoint gradient trajectory & $\cos(\nabla I_X,\nabla I_T)\approx0$; update target-aligned only at init & PASS \\
H2 & Learned weak-alignment trajectory & training checkpoints & $\rho$(init)$=0.71/0.72 \to \rho$(10 epochs)$=0.39/0.05$; local ranks stabilize earlier & PASS \\
H3 & Axis-specific propagation & canonical 3-backbone suite & Local$\to$local, Target$\to$target $\bar\rho > 0.1$; cross $\approx 0$ & PASS \\
H4 & C5: Best local $>$ best target & fixed pruning suite & $\Delta$AUC: +8.0 / +6.6 / +10.1pp (FLOPs) & PASS \\
H4 & C6: Best local $>$ Taylor (FLOPs) & fixed pruning suite & $\Delta$AUC: +37.2 / +1.1 / +12.2pp (FLOPs) & PASS \\
H4 & C7: Hybrid alloc/rank is architecture-dependent & fixed pruning suite & VGG: +2.0pp; ResNet: $-14.0$pp; MobileNet: $-0.1$pp & PASS \\
H4 & C8: Beyond-norm gain is architecture-dependent & weight sweep & VGG best Mag$+0.75 I_X$; ResNet/Mobile still favor local scores & PASS \\
H4 & C9: Random/FPGM/BN controls & additional-control suite & local $>$ target and CHIP on 3/3; VGG norm-like baselines remain strongest & PASS$^\dagger$ \\
H4 & Matched-task lesion control & peer-replacement suite & peer $R^2$ predicts recovery at matched task MI; $I_X$/task MI near zero & PASS \\
\bottomrule
\end{tabular}
\\[4pt]
\raggedright\footnotesize All pruning claims use global-threshold pruning with FLOPs-matched evaluation under the unified fixed pruning suite.
C2b: Supports target-side collapse as a broad structural statement; it does not imply strong channelwise estimator agreement.
C5: Appendix~\ref{app:benchmark_pruning_extension} extends this comparison to 11 added cells spanning CIFAR-10, Tiny-ImageNet, and ImageNet-100, where the best local method still exceeds the best target method in all 11 cells and CHIP in the same 11 cells.
C6: The strongest local method exceeds Taylor on all three canonical architectures, but the margin is small on VGG-16; in the breadth suite, local still exceeds Taylor in 9 of the 11 added cells, while Taylor remains stronger on CIFAR-10/ResNet-18 and ImageNet-100/AlexNet.
C7: Hybrid allocation/ranking is no longer uniformly beneficial under the fixed protocol; only VGG-16 shows a positive gain over the best non-hybrid method.
C8: A targeted uniform-allocation weight sweep shows that adding local $I_X$ to magnitude helps on VGG-16, but not enough to displace the strongest pure local scores on ResNet-18 or MobileNetV2.
C9: Table~\ref{tab:topup_controls} is an expanded-control comparison over a stricter common FLOPs range; it supports local-vs-target and local-vs-CHIP, while explicitly marking VGG-16 as a classical-baseline boundary case.
\end{table}

\paragraph{A note on clustering quality vs.\ pruning utility.}
Clustering quality (as measured by silhouette or geometric separation) and pruning utility (accuracy retention under channel removal) need not align.
PID-space clusters may be geometrically cleaner, but local replaceability metrics are more intervention-aligned for channel removal.
This distinction between descriptive structure and causal relevance for a downstream task is central to our two-axis framework.

\section{Benchmark Consistency Across Datasets and Backbones}
\label{app:benchmark_consistency}

\begin{figure*}[t]
    \centering
    \includegraphics[width=0.98\textwidth]{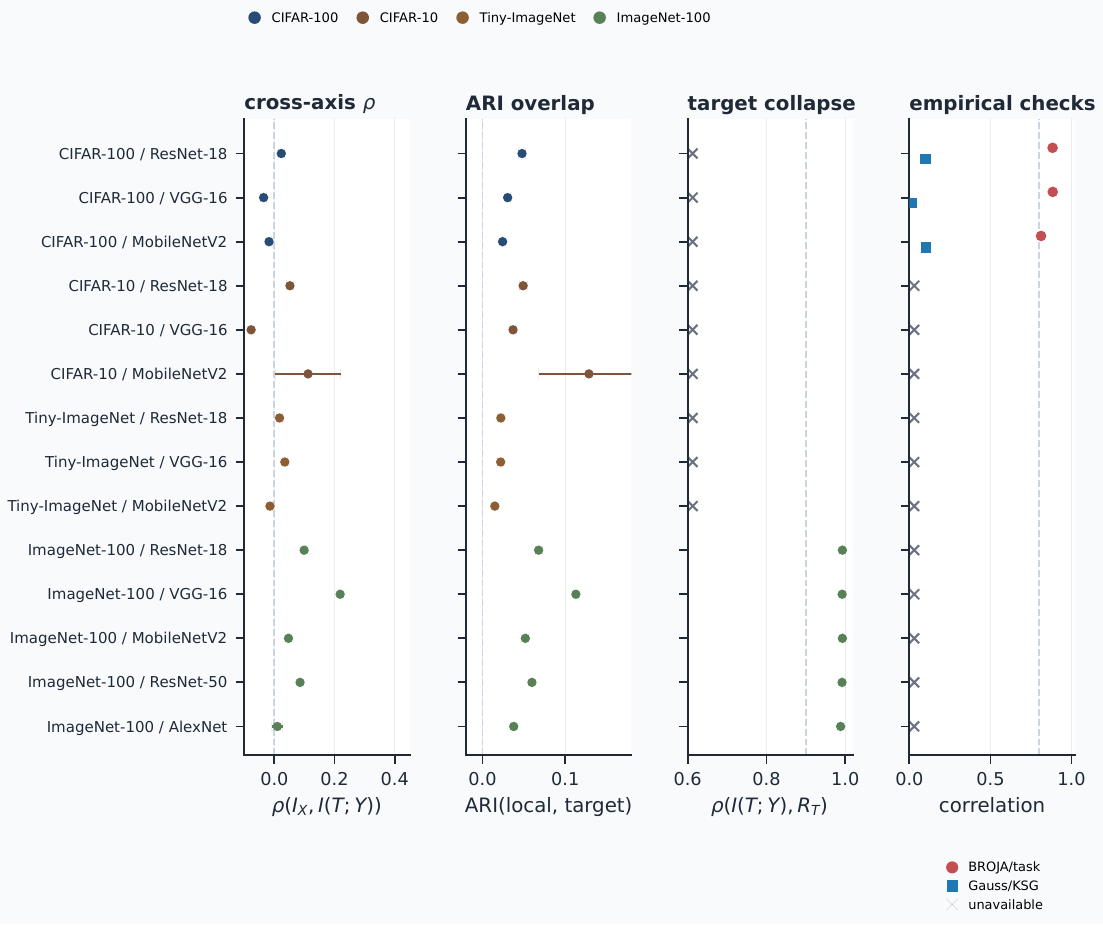}
    \caption{\textbf{Broader benchmark consistency for the structural claims.}
    Each row is one dataset/backbone cell, with points showing mean$\pm$SEM over reusable checkpoints from that family.
    (A) Cross-axis correlation $\rho(I_X, I(T;Y))$.
    (B) ARI between local $(I_X,\bar{R}_X)$ and target $(I(T;Y),\Syn)$ clusterings.
    (C) Gaussian target-side collapse, measured by $\rho(I(T;Y), R_T)$, on the newer ImageNet-100 families that retain target PID fields.
    (D) Canonical CIFAR-100 empirical target checks: BROJA/task is the correlation between BROJA shared information and task MI; Gauss/KSG is the rank agreement between Gaussian and KSG task-MI estimates; crosses mark unavailable runs.
    The exact magnitudes vary by benchmark cell, but the overall picture is stable: cross-axis alignment stays far from collapse, while target-side collapse remains extremely strong where it can be measured directly.}
    \label{fig:app_benchmark_consistency}
\end{figure*}

The main text keeps CIFAR-100 as the visualization domain because it gives the cleanest mechanistic comparisons.
To test whether the qualitative direction of the structural claims survives beyond that setting, we aggregated 68 raw runs spanning CIFAR-100, CIFAR-10, Tiny-ImageNet, and ImageNet-100 across ResNet-18, VGG-16, MobileNetV2, ResNet-50, and AlexNet.
Figure~\ref{fig:app_benchmark_consistency} reports a compact benchmark dashboard rather than a second full figure suite.
We keep the ConvNeXt pilot separate from Figure~\ref{fig:two_axes} and from this pooled dashboard because it is meant as a modern-CNN stress test, and its partially different H1 behavior would otherwise blur the canonical CIFAR-100 visualization anchor.

Two points matter.
First, H1 remains weakly aligned rather than collapsed across all benchmark cells: mean cross-axis correlation $\rho(I_X, I(T;Y))$ stays small overall (grand mean $0.039$), and mean ARI between local and target clusterings remains low (grand mean $0.050$), even though the exact magnitude varies by dataset and backbone.
Second, where the newer raw runs store the Gaussian target PID fields, the target-collapse component of H1 remains strong: on the ImageNet-100 family, $\rho(I(T;Y), R_T)$ stays between $0.988$ and $0.993$ across ResNet-18, VGG-16, MobileNetV2, ResNet-50, and AlexNet.
Empirical target-side checks are still only available for the canonical CIFAR-100 three-backbone suite, so this section should be read as a broader consistency check for direction, not as a replacement for the main CIFAR-100 evidence base.

\section{Fixed-Protocol Pruning Breadth Across Extra Benchmark Cells}
\label{app:benchmark_pruning_extension}

To complement the broader structural benchmark layer, we ran a broader fixed-protocol pruning suite on 11 additional benchmark cells: CIFAR-10 with ResNet-18, VGG-16, and MobileNetV2; Tiny-ImageNet with the same three backbones; and ImageNet-100 with ResNet-18, VGG-16, MobileNetV2, ResNet-50, and AlexNet.
These runs reuse trained checkpoints and apply the same 200-batch-capped pruning protocol used in the main paper.
The goal is narrow: check whether the intervention ordering survives outside the canonical CIFAR-100 three-backbone suite, not to define a second ranking benchmark.

The cleanest result is that the best local method still exceeds the best target method in all 11 added cells, with a mean $\Delta$AUC of $+0.080$ across the breadth suite.
This remains true against the dependency-aware CHIP baseline: the best local method exceeds CHIP in all 11 added cells, with mean $\Delta$AUC $+0.133$.
The beyond-norm picture is also consistent with the current paper boundary: the best mixed norm-plus-local score exceeds pure magnitude in 10 of the 11 added cells and ties it on CIFAR-10/ResNet-18, with mean $\Delta$AUC $+0.023$.
By contrast, local does \emph{not} uniformly dominate Taylor outside the canonical suite: Taylor remains stronger on CIFAR-10/ResNet-18 by $0.055$ AUC and on ImageNet-100/AlexNet by $0.002$ AUC, while the best local method wins on the other nine added cells.
We therefore use this section to strengthen the H4 ranking distinction, to show that the beyond-norm story remains conditional rather than universal, and to document that the local-family advantage persists even against a dependency-aware baseline.

\paragraph{Practical local-family selector.}
To test whether the architecture-dependent local recipe requires oracle access, we ran a leave-one-seed-out selector within each dataset/backbone cell.
For each held-out seed, the selector chooses the local-axis score with higher mean AUC on the other seeds, then evaluates that fixed choice on the held-out seed.
This simple selector remains positive against the best target score on 52/53 held-out seeds (mean $\Delta$AUC $+0.076$), positive against CHIP on 53/53 (mean $+0.121$), and positive against Taylor on 43/53 (mean $+0.040$), with mean gap to oracle best-local of only $-0.001$ AUC.
Thus the actionable recommendation does not require knowing the winning local score in advance: choose the local family, then calibrate within it using a small validation sweep.

\begin{table}[t]
\centering
\small
\begin{tabular}{lcc}
\toprule
Comparison & Positive cells & Mean $\Delta$AUC \\
\midrule
Best local $-$ best target & 11 / 11 & $+0.080$ \\
Best local $-$ CHIP & 11 / 11 & $+0.133$ \\
Best local $-$ Taylor & 9 / 11 & $+0.044$ \\
Best mixed $-$ magnitude & 10 / 11 & $+0.023$ \\
\bottomrule
\end{tabular}
\caption{\textbf{Breadth-suite sign summary.}
Compact view of the fixed-protocol pruning breadth controls over the 11 added benchmark cells.
The stable metric-family result is that local ranking beats both target ranking and CHIP across the full breadth suite, while the Taylor and beyond-norm comparisons remain conditional.}
\label{tab:benchmark_pruning_breadth_summary}
\end{table}

\begin{figure*}[t]
    \centering
    \includegraphics[width=0.98\textwidth]{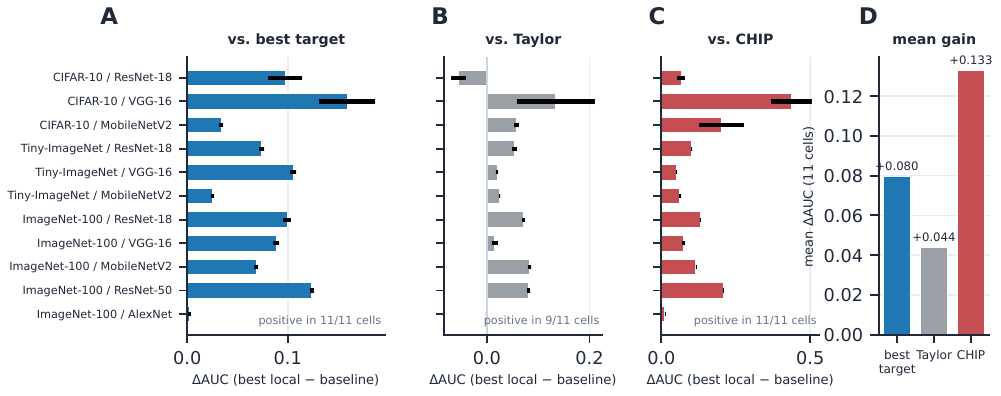}
    \caption{\textbf{Fixed-protocol pruning breadth: best-local vs.\ prior baselines across 11 benchmark cells.}
    Each row in panels A--C is one dataset/backbone cell from the fixed breadth suite (CIFAR-10 $\times$ 3 backbones, Tiny-ImageNet $\times$ 3, ImageNet-100 $\times$ 5); bars are mean~$\pm$~SEM over 3--5 reusable checkpoints per cell.
    (A) vs.\ best target PID score, positive on 11/11 cells.
    (B) vs.\ Taylor, positive on 9/11 cells (negative on CIFAR-10/ResNet-18 and near zero on ImageNet-100/AlexNet).
    (C) vs.\ CHIP~\citep{sui2021chip}, positive on 11/11 cells.
    (D) Mean gain over each baseline, pooled across the 11 cells.
    Best-local methods beat the dependency-aware information-theoretic baseline (CHIP) by a large margin (+0.133 AUC on average), and beat the best single-score target method by +0.080; the weaker Taylor margin is consistent with Taylor being partly local-axis-sensitive (\S\ref{sec:learned}).}
    \label{fig:app_benchmark_pruning_extension}
\end{figure*}

Table~\ref{tab:benchmark_pruning_breadth_summary} gives the compact sign summary, and Figure~\ref{fig:app_benchmark_pruning_extension} shows the same breadth-suite comparisons cell by cell.

\section{Modern CNN Stress Test: ConvNeXt-T / ImageNet-100}
\label{app:convnext_pilot}

To test whether the picture extends beyond the older CNN families without changing the unit of analysis, we ran a pilot on ConvNeXt-T with ImageNet-100 (3 seeds).
We treat this as a stress test rather than as primary evidence: ConvNeXt remains channel-based and fits the same Conv2d analysis pipeline, but it is architecturally more modern than the canonical ResNet/VGG/MobileNet suite.
We therefore report it separately rather than replacing VGG-16 in the canonical figures: VGG-16 remains important for the paper because it is the clearest case where magnitude is unusually strong and the beyond-norm gain is architecture-dependent.

The pilot gives a useful but nuanced picture.
The target-side collapse component of H1 remains extremely strong: the Gaussian collapse still holds, with mean $\rho(I(T;Y), \Red_T) = 0.995 \pm 0.0002$ across the ConvNeXt layer summaries.
H4 also survives cleanly: under the same pruning protocol, the best local method exceeds the best target method in all 3 seeds, with mean $\Delta$AUC $+0.107 \pm 0.005$.
The best local method is consistently CAP-QGW, the best target method is consistently PID-no-red, and the local margin over Taylor is small but still positive on average ($+0.006 \pm 0.004$ AUC), while the local margin over pure magnitude remains large ($+0.179 \pm 0.008$ AUC).

The H1 separation is weaker on ConvNeXt.
The cross-axis alignment is still far from collapse, but it is more coupled than in the older CNN benchmark layer: mean $\rho(I_X, I(T;Y)) = 0.154 \pm 0.011$ and mean ARI between local and target clusterings is $0.108 \pm 0.005$.
We therefore use ConvNeXt as a supportive modern-CNN control, not as evidence that every part of the CIFAR-100 CNN picture transfers unchanged.
Table~\ref{tab:convnext_pilot} summarizes the pilot.

\begin{table}[t]
\centering
\small
\begin{tabular}{lc}
\toprule
ConvNeXt-T / ImageNet-100 pilot (3 seeds) & Mean $\pm$ SEM \\
\midrule
Cross-axis $\rho(I_X, I(T;Y))$ & $0.154 \pm 0.011$ \\
ARI(local, target) & $0.108 \pm 0.005$ \\
$\rho(I(T;Y), \Red_T)$ & $0.995 \pm 0.0002$ \\
Best local $-$ best target AUC & $+0.107 \pm 0.005$ \\
Best local $-$ Taylor AUC & $+0.006 \pm 0.004$ \\
Best local $-$ magnitude AUC & $+0.179 \pm 0.008$ \\
\bottomrule
\end{tabular}
\caption{\textbf{Appendix-only ConvNeXt-T/ImageNet-100 stress test.}
This pilot extends the paper to a more modern CNN backbone without changing the unit of analysis.
The target-side collapse and pruning hierarchy survive, while the within-layer axis separation is weaker than in the older CNN families.}
\label{tab:convnext_pilot}
\end{table}

\section{Uniform-Allocation Weight Sweep}
\label{app:metric_weight_sweep}

\begin{figure}[tbp]
    \centering
    \includegraphics[width=0.98\columnwidth]{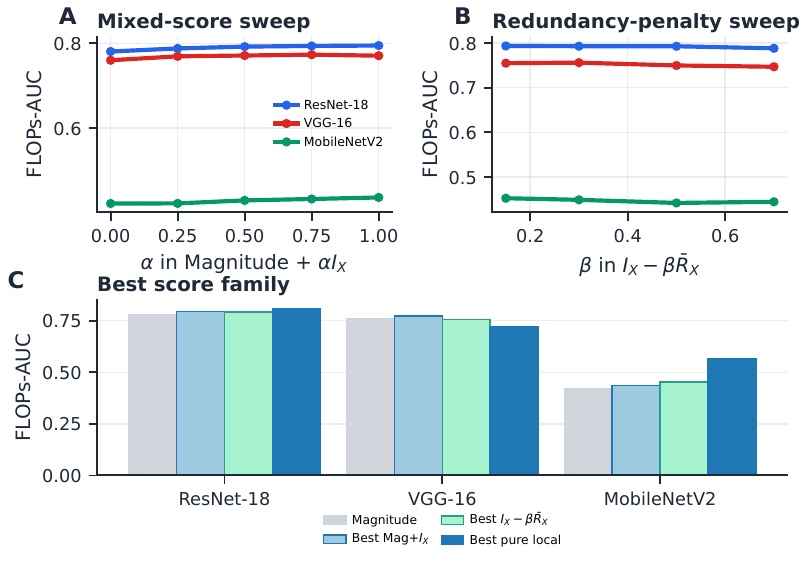}
\caption{\textbf{Targeted uniform-allocation weight sweep.}
    Uniform allocation only: this figure isolates score construction from cross-layer allocation and should not be compared directly to the global-threshold AUC values in Figure~\ref{fig:pruning_intervention}.
    (A) FLOPs-AUC as the mixed score $\text{Magnitude} + \alpha I_X$ varies.
    VGG-16 improves monotonically up to $\alpha \approx 0.75$, while ResNet-18 shows only a modest gain and MobileNetV2 remains far below the strongest pure local score.
    (B) FLOPs-AUC for $I_X - \beta \bar{R}_X$.
    Small redundancy penalties help mildly on ResNet-18 and VGG-16 but do not beat the best local composite on ResNet-18 or MobileNetV2.
    (C) Best-of-family comparison.
    Composite-$I_X$ remains best on ResNet-18 and MobileNetV2, while a mixed norm-plus-local score is best on VGG-16.}
    \label{fig:app_metric_weight_sweep}
\end{figure}

To isolate \emph{score construction} from \emph{cross-layer allocation}, we ran a targeted uniform-allocation sweep on three local families: pure magnitude, mixed magnitude-plus-$I_X$ scores, and $I_X$ with a small redundancy penalty.
This sweep clarifies the strongest ``beyond norm'' claim supported by the current data: adding local input-capture information to magnitude helps on VGG-16, but the best score family remains architecture-dependent.
It should be read as the detailed appendix version of H4, not as a new standalone pruning claim.
Figure~\ref{fig:app_metric_weight_sweep} reports the sweep.

The main paper uses this only to clarify the boundary.
It supports a conditional H4: the local pruning advantage is not reducible to norm alone, but the best way to combine norm and local information depends on architecture.
Table~\ref{tab:norm_local_boundary} gives the corresponding norm/local diagnostic.
Weight norm is moderately aligned with $I_X$ in all three backbones and is weakly or negatively aligned with peer overlap, so the VGG-16 magnitude boundary should not be read as evidence that local replaceability collapses to norm.
The component ablation is consistent with this: over 130{,}400 channel rows from the same three backbones, within-layer standardized lesion-damage $R^2$ rises from $0.0247$ for norm alone to $0.0379$ for $(I_X,\bar R_X)$, and at matched $I_X$, higher peer overlap gives lower damage in 55.3\% of matched pairs ($p<10^{-100}$).

\begin{table}[htbp]
\centering
\caption{\textbf{Norm/local boundary diagnostic.}
Within-layer Spearman correlations and SD summaries over the locked CIFAR-100 layer--seed cells.
Norm partially tracks the input-capture coordinate, but it does not encode peer overlap $\bar R_X$.}
\label{tab:norm_local_boundary}
\small
\resizebox{\linewidth}{!}{\begin{tabular}{lcccc}
\toprule
Backbone & $\rho(\log\|w\|^2,I_X)$ & $\rho(\log\|w\|^2,\log\RQ)$ & $\rho(\log\|w\|^2,\bar R_X)$ & SD$(\log\RQ)$ \\
\midrule
ResNet-18 & $0.641 \pm 0.020$ & $-0.076 \pm 0.024$ & $-0.119 \pm 0.024$ & $0.334 \pm 0.019$ \\
VGG-16 & $0.581 \pm 0.034$ & $-0.157 \pm 0.043$ & $-0.228 \pm 0.023$ & $0.396 \pm 0.032$ \\
MobileNetV2 & $0.559 \pm 0.013$ & $0.024 \pm 0.019$ & $0.054 \pm 0.018$ & $0.694 \pm 0.051$ \\
\bottomrule
\end{tabular}
}
\end{table}

\section{Lesion and Pairwise Replaceability Details}
\label{app:pairwise}

This section gives the detailed peer-structure and lesion analyses behind the main replaceability claims.
It starts with within-layer pairwise redundancy matrices, then tests whether compact peer support predicts actual removal and replacement behavior.
For each layer, we compute the full $N \times N$ pairwise redundancy matrix $R_{ij} = -\frac{1}{2}\log(1 - \rho_{ij}^2)$ between all channel pairs, then reorder channels by their descriptive local cluster labels.

\begin{figure}[tbp]
    \centering
    \includegraphics[width=0.98\columnwidth]{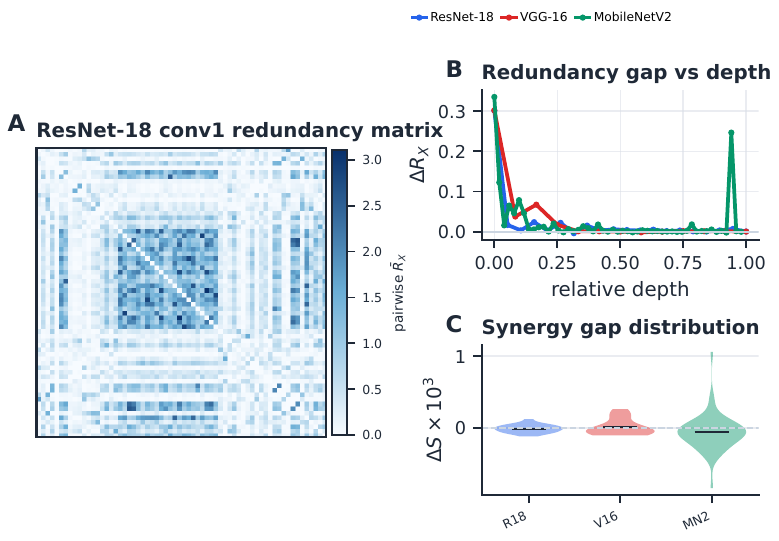}
    \caption{\textbf{Pairwise redundancy dominates pairwise organization.}
    (A) Pairwise redundancy matrix for ResNet-18 conv1, with channels reordered by descriptive local cluster label.
    Clear block-diagonal structure: within-type pairwise redundancy exceeds between-type redundancy, particularly in early layers.
    (B) Within-type minus between-type pairwise redundancy ($R_\text{within} - R_\text{between}$) vs.\ relative depth.
    The difference is largest in early layers and decays with depth, consistent with decreasing channel specialization.
    (C) Distribution of the within-type minus between-type pairwise synergy gap across layers.
    Unlike redundancy, the synergy gap remains centered near zero, indicating that the cluster structure is defined primarily by shared information rather than pairwise combinatorial effects.}
    \label{fig:pairwise}
\end{figure}

Key findings: (1) Within-type pairwise redundancy exceeds between-type redundancy by a factor of 1.14--1.25 in early layers (Cohen's $d \approx 0.3$--$0.8$), confirming that the descriptive groupings capture genuine overlap structure.
(2) This within/between gap decreases with depth, consistent with layers becoming less specialized.
(3) Pairwise synergy stays near its within/between null level, indicating that clusters are defined primarily by redundancy (shared information), not by synergy (combinatorial contributions).
Figure~\ref{fig:pairwise} summarizes these pairwise controls.

\subsection{R-graph vs S-graph: different topologies}

Treating each pairwise matrix as a weighted graph and keeping the top 10\% of edges yields two graphs per layer.
Greedy Newman modularity~\citep{newman2006modularity} over all conv layers of three backbones (5 seeds each) gives a consistent depth profile: the R-graph is uniformly more modular than the S-graph at every relative depth on every backbone, with the gap largest in input-side layers and shrinking toward the output (Figure~\ref{fig:higher_order_depth}A).
Redundant channels cluster into more clearly-modular groups than synergistic pairs throughout the network depth.

\subsection{Replaceability hulls}
\label{app:hulls}

\begin{figure*}[t]
\centering
\includegraphics[width=0.98\textwidth]{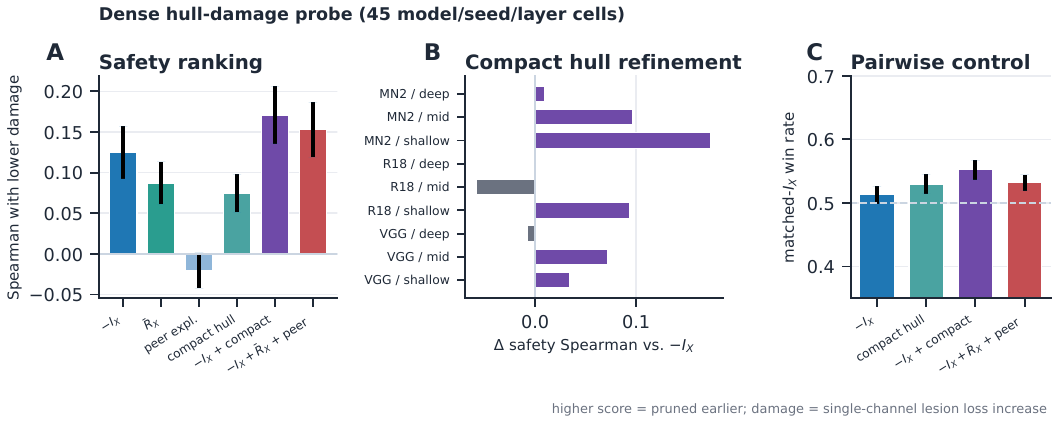}
\caption{\textbf{Direct lesion evidence for replaceability.}
Single-channel ablations without fine-tuning on the fixed CIFAR-100 evaluation split.
All scores are oriented so that larger values predict lower lesion damage; \emph{peer} denotes $\bar R_X$, \emph{compact hull expl.} denotes $E_i^\mathrm{full}/\max(1,|H_i|)$, and ``+'' labels add standardized support terms to $-I_X$.
(A) Removal ranking; positive Spearman means agreement with lower damage.
(B) Gain from compact peer support over $-I_X$.
(C) Matched-$I_X$ control. Compact peer support gives a modest local-axis refinement.}
\label{fig:hull_damage_dense}
\end{figure*}

For each channel $i$ we greedily select the minimum peer set $S_i$ whose conditional Gaussian explanation of $c_i$ is within $\varepsilon=5\%$ of the full-peer explanation, capped at $|S|\leq 10$.
Depth profiles across 3 models $\times$ 5 seeds (Figure~\ref{fig:higher_order_depth}B,C) resolve the local axis into two depth-separated regimes.
At input-side layers 28--48\% of channels are \emph{singleton-hull}: each has a single near-duplicate in the same layer.
At the deepest convs 67--78\% saturate the greedy cap ($|S_i|\geq 10$), requiring a fully-distributed peer set.
Mean hull size grows monotonically from $\sim\!1.5$ to $\sim\!7$--$8$ with depth.
This sharpens the pairwise-$R$ summary: input layers are not just ``redundant on average'' but \emph{paired}, while deep layers carry genuinely-distributed roles.
We therefore use hulls as a structural descriptor of the local axis.
A dense hull-damage probe tests whether this higher-order structure improves channel-removal prediction.
Across 45 model/seed/layer cells (3 backbones $\times$ 5 seeds $\times$ 3 depth slices), raw hull size is \emph{not} a useful pruning score: higher hull size correlates with \emph{higher}, not lower, lesion damage (mean matched-$I_X$ win rate $0.474$).
The useful refinement is instead \emph{compact peer explanation}: combining low $I_X$ with concentrated peer support improves removal ranking over plain $-I_X$ (Spearman with lower lesion damage $0.171 \pm 0.036$ vs.\ $0.125 \pm 0.032$; matched-$I_X$ win rate $0.552 \pm 0.015$ vs.\ $0.513 \pm 0.014$).
We also run an explicit peer-replacement control: for each sampled channel, fit a ridge-linear reconstruction from the top-8 same-layer peers on the calibration split and replace the channel activation at evaluation time.
Because channels with near-zero or beneficial lesion damage make recovered-fraction ratios unstable, Table~\ref{tab:peer_replacement_control} reports robust channel-level summaries after requiring a minimum positive lesion damage.
When the lesion is nontrivial, peer replacement usually recovers most of the damage, and reconstruction $R^2$ is the strongest recovery predictor among the tested scalar quantities.
Thus hulls help as a mechanism diagnostic and local-score refinement, but we do not promote hull size as a new main pruning metric.
Figure~\ref{fig:hull_damage_dense} reports the dense hull-damage probe behind the direct lesion summary in the main text.

\begin{table}[htbp]
\centering
\caption{\textbf{Peer-replacement intervention control} (45 model/seed/layer cells; 2720 sampled channels).
Rows restrict to channels whose lesion increases loss by more than the stated threshold. Recovery is $(\Delta L_\mathrm{lesion}-\Delta L_\mathrm{replace})/\Delta L_\mathrm{lesion}$; peer $R^2$ is the top-8 linear reconstruction fit.}
\label{tab:peer_replacement_control}
\small
\begin{tabular}{@{}lcccc@{}}
\toprule
Lesion threshold & Median recovery & Peer helps & $\rho$(peer $R^2$, recovery) & $\rho$(task MI, recovery) \\
\midrule
$10^{-4}$ & $0.666$ & $0.808$ & $0.357$ & $0.176$ \\
$10^{-3}$ & $0.687$ & $0.843$ & $0.423$ & $0.225$ \\
$5{\times}10^{-3}$ & $0.799$ & $0.899$ & $0.516$ & $0.284$ \\
\bottomrule
\end{tabular}
\end{table}

\begin{table}[htbp]
\centering
\caption{\textbf{Matched-task lesion/replacement control.}
Channels are binned by task MI within each model/seed/layer cell before residualizing the tested predictors.
Peer-reconstruction $R^2$ remains predictive of recovery at matched task relevance, whereas $I_X$ and task MI do not.
Intervals are bootstrap CIs for Spearman correlations and Wilson CIs for high-peer win rates.}
\label{tab:matched_task_lesion_control}
\small
\resizebox{\linewidth}{!}{\begin{tabular}{lccccc}
\toprule
Threshold & Channels & peer $R^2$ $\rho$ & $I_X$ $\rho$ & task MI $\rho$ & high-peer win \\
\midrule
$10^{-4}$ & 1409 & $0.131$ [0.077, 0.183] & $-0.012$ [-0.060, 0.038] & $0.038$ [-0.013, 0.088] & $0.603$ [0.537, 0.665] \\
$10^{-3}$ & 1069 & $0.218$ [0.152, 0.282] & $0.006$ [-0.057, 0.068] & $-0.004$ [-0.064, 0.057] & $0.630$ [0.549, 0.704] \\
$5{\times}10^{-3}$ & 624 & $0.258$ [0.167, 0.347] & $-0.002$ [-0.084, 0.084] & $-0.048$ [-0.131, 0.035] & $0.645$ [0.544, 0.735] \\
\bottomrule
\end{tabular}
}
\end{table}

\subsection{Triplet target-excess}
\label{app:triplets}

We compute triplet target-excess over the best pair, $S_3(T; c_i,c_j,c_k) = I(T;[c_i,c_j,c_k]) - \max_\text{pair} I(T;[c_a,c_b])$, on the top-24 task-relevant channels per layer, across 3 backbones $\times$ 5 seeds (up to 1500 sampled triples per layer).
The $S_3/S_2$ ratio climbs from $\sim\!0.2$--$0.3$ at the input to $\sim\!0.4$--$0.55$ at the output on all three models (Figure~\ref{fig:higher_order_depth}D), most steeply on VGG-16.
Deep-layer task content increasingly has excess information at the triplet level beyond the best pair.
This is complementary to the $\mathrm{SI}/\mathrm{CI}$ U-shape in \S\ref{sec:beyond_mmi}: one characterises the ratio of shared to synergistic pair content, the other the order of the synergistic structure.

\subsection{Higher-order structure across training}
\label{app:higher_order_training}

\begin{figure*}[tbp]
    \centering
    \includegraphics[width=0.98\textwidth]{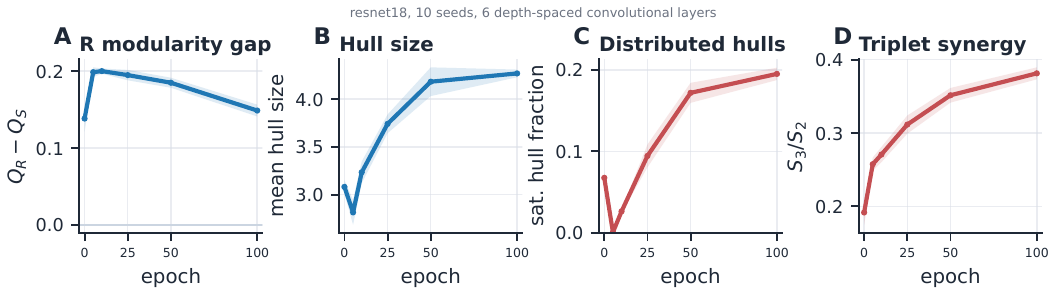}
    \caption{\textbf{Higher-order structure across training.}
    ResNet-18/CIFAR-100, 10 checkpointed seeds, six depth-spaced convolutional layers, mean $\pm$ SEM over seeds after averaging layers within seed.
    $Q_R-Q_S$ is the local-redundancy minus target-excess graph modularity gap; ``sat.'' denotes the saturated-hull fraction ($|H_i|\geq10$).
    The R-vs-S modularity gap is already positive at initialization, but hull size, distributed-hull fraction, and triplet target-excess $S_3/S_2$ grow over training.}
    \label{fig:higher_order_training}
\end{figure*}

To test whether the higher-order patterns are fixed by architecture or sharpen during learning, we analyzed 10 checkpointed ResNet-18 trajectories at epochs 0, 5, 10, 25, 50, and 100.
For each seed and epoch, we recomputed the R-graph/S-graph modularity gap, replaceability hull statistics, and triplet target-excess on six depth-spaced convolutional layers using a fixed 2000-image CIFAR-100 calibration subset and the same Gaussian proxies as the main analysis.
The result is not simply ``absent at initialization, present after training.''
The R-graph is already more modular than the S-graph at initialization ($Q_R-Q_S=0.138\pm0.017$) and remains so at convergence ($0.149\pm0.008$), with a larger early-training peak near epoch 10.
By contrast, the more distributed higher-order quantities grow with learning: mean hull size increases from $3.08\pm0.07$ to $4.27\pm0.04$, the saturated-hull fraction from $0.067\pm0.006$ to $0.195\pm0.008$, and $S_3/S_2$ from $0.192\pm0.017$ to $0.381\pm0.008$.
Thus some local-topology bias is present before training, but learning makes replaceability more distributed and target information more triplet-level.
Figure~\ref{fig:higher_order_training} shows these trajectories.

\section{Cross-Layer Propagation Definition}
\label{app:crosslayer_extra}

This appendix gives the formal definition behind the cross-layer propagation matrices reported in Figure~\ref{fig:learned_main}E--H.
For each destination channel $j$ in layer $\ell{+}1$, we compute the weight-routed source summary
\begin{equation}
    \bar{m}_s(j) = \frac{\sum_{i} \|W_{ji}\|_F \cdot m_s^{(\ell)}(i)}{\sum_{i} \|W_{ji}\|_F},
    \label{eq:weighted_metric}
\end{equation}
where $W_{ji}$ is the weight kernel connecting source channel $i$ to destination channel $j$, and $m_s^{(\ell)}(i)$ is a metric of source channel $i$ at layer $\ell$.
Using Eq.~\ref{eq:weighted_metric}, we compute $\rho(\bar{m}_s, m_d^{(\ell+1)})$ for all source--destination metric pairs $m_s, m_d \in \{I_X, \bar{R}_X, I(T;Y), \Syn\}$.
The resulting propagation matrix is block-diagonal in the two axes: within-axis correlations (local$\to$local $\bar{\rho}=0.12$--$0.23$; target$\to$target $\bar{\rho}=0.09$--$0.21$) are modest but above cross-axis correlations ($\bar{\rho}\approx0$).
The per-architecture matrices and the within- vs cross-axis summary are plotted in Figure~\ref{fig:learned_main}E--H.

\section{One-Dimensional vs Two-Axis Ranking Example}
\label{app:teaser}

\begin{figure}[tbp]
    \centering
    \includegraphics[width=0.98\textwidth]{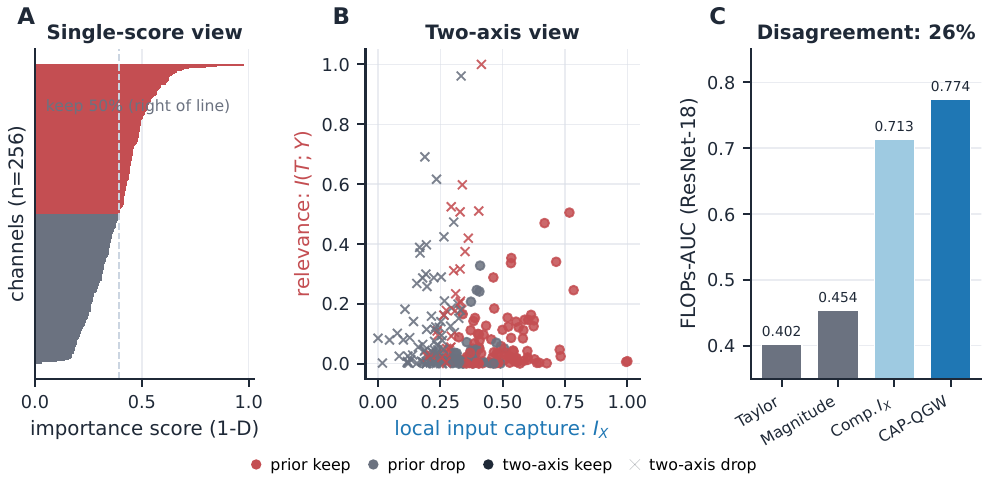}
    \caption{\textbf{Why two axes matter for pruning.}
    (A) A single-score importance view of one conv layer ($n=256$ channels of ResNet-18 layer3.0.conv2, seed 42): channels are ranked along a 1-D axis and the top 50\% are kept.
    (B) The same channels plotted in the two-axis plane $(I_X, I(T;Y))$: red = kept under the prior view, grey = dropped. Markers show the two-axis decision ($\circ$ kept / $\times$ dropped). About 26\% of channels disagree, and most disagreement lies along the relevance axis $I(T;Y)$, which a one-dimensional score cannot resolve.
    (C) Locked-suite FLOPs-AUC on ResNet-18/CIFAR-100 (Table~\ref{tab:pruning_flops_appendix}): local-axis scores Composite-$I_X$ and CAP-QGW retain substantially more accuracy under FLOPs-matched recovery than Taylor or magnitude.}
    \label{fig:teaser}
\end{figure}

Figure~\ref{fig:teaser} visualizes the introductory 1-D versus two-axis mismatch.

\section{Target-Side BROJA Decomposition}
\label{app:theory_figure}

\begin{figure}[tbp]
    \centering
    \includegraphics[width=0.85\textwidth]{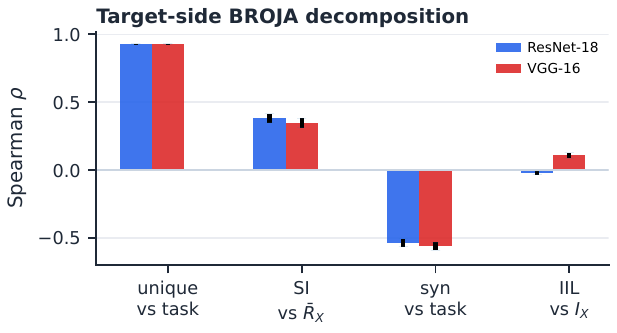}
    \caption{\textbf{BROJA target-side PID decomposition} (pooled within-layer Spearman correlations on ResNet-18 and VGG-16; mean $\pm$ SEM over layers).
    BROJA unique information aligns with task MI ($\bar\rho\approx0.93$); BROJA synergy anti-aligns with task MI ($\bar\rho\approx-0.55$); BROJA shared information SI tracks $\bar R_X$ only weakly ($\bar\rho\approx0.35$); and the irrecoverable information loss IIL (defined above) is near zero against $I_X$.
    BROJA SI is therefore a target-directed quantity, not a relabeling of local-axis overlap.}
    \label{fig:broja_decomposition}
\end{figure}

Section~\ref{sec:beyond_mmi} reports the non-parametric BROJA target-side check.
Figure~\ref{fig:broja_decomposition} visualizes the corresponding atom-level structure on the canonical CIFAR-100 suite, beyond the scalar correlations quoted in the main text.
We use the standard BROJA atoms shared (SI), unique, and synergy from the bivariate decomposition $I(T;[Y_i,Y_j]) = \mathrm{SI} + \mathrm{Uniq}_i + \mathrm{Uniq}_j + \mathrm{Syn}$~\citep{bertschinger2014quantifying}, plus a derived quantity we call the \emph{irrecoverable information loss} (IIL): for each channel $i$ relative to its top-MI peer $j^\star$, $\mathrm{IIL}(i) = \mathrm{Uniq}_{i\mid j^\star}$ is the per-channel BROJA-unique mass not recoverable from the best peer, normalized to the channel's task MI.
IIL is the BROJA-side counterpart of peer-replacement loss for a single peer; we plot only its rank correlation with $I_X$ as a sanity check that local input capture and irrecoverable target uniqueness are different orderings.

\section{Pilot Extensions}
\label{app:rule_experiments}

This appendix gathers two pilot experiments that extend the two-axis interpretation beyond the main pruning analyses: a local-axis decorrelation regularizer (used during training) and a frozen-ViT transfer test (used post hoc on a non-convolutional backbone).
They are not part of the primary H1--H4 evidence chain, but they show that the local axis can be acted on during training and that the weak-alignment picture is not specific to convolutional networks.

\paragraph{Decorrelation regularizer.}
We train ResNet-18 on CIFAR-100 with the standard SGD recipe (lr $=0.1$, momentum $0.9$, weight decay $5 \cdot 10^{-4}$, cosine schedule, 100 epochs), adding a regularizer
\[
\mathcal{L}_\text{reg} \;=\; \lambda \;\frac{1}{|\mathcal{S}|}\sum_{\ell \in \mathcal{S}}\bar R_X^{(\ell)},
\]
where $\mathcal{S}$ is a depth-spaced subset of at most 6 \texttt{Conv2d} modules, $\bar R_X^{(\ell)}$ is the mean Gaussian shared-information summary over up to 128 subsampled channels of layer $\ell$ (64 reference channels), and the regularizer is applied every second mini-batch to keep the per-step overhead bounded.
Table~\ref{tab:rule3_acc} reports top-1 accuracy and the measured post-hoc $\bar R_X$ on all three trained networks, using a common 2000-image calibration split and the same depth-spaced subset of 6 \texttt{Conv2d} modules for every $\lambda$.
The baseline ($\lambda=0$) network carries mean $\bar R_X = 0.103$; at $\lambda=0.1$ this drops to $0.034$ while accuracy moves from $0.7736$ to $0.7748$.

\begin{table}[htbp]
\centering
\small
\begin{tabular}{lccr}
\toprule
$\lambda$ & top-1 val acc & mean $\bar R_X$ (post-hoc) & $\Delta\bar R_X$ vs.\ baseline \\
\midrule
$0.00$ (baseline) & $0.7736$ & $0.1033$ & n/a \\
$0.01$            & $0.7697$ & $0.0510$ & $-51\%$ \\
$0.10$            & $\mathbf{0.7748}$ & $\mathbf{0.0341}$ & $\mathbf{-67\%}$ \\
\bottomrule
\end{tabular}
\caption{\textbf{Decorrelation regularizer on ResNet-18/CIFAR-100, seed 42, 100 epochs.}
$\bar R_X$ is measured post-hoc by replaying the same hook protocol used during training on a fixed 2000-image calibration split, so the $\lambda=0$ value is directly comparable to the regularized networks.
At $\lambda=0.1$ redundancy drops by $67\%$ relative to baseline while top-1 accuracy slightly increases.}
\label{tab:rule3_acc}
\end{table}

\paragraph{ViT-B/16 transfer test.}
We load torchvision's ViT-B/16~\citep{dosovitskiy2021image} with ImageNet-1k weights (no fine-tuning), pass $N=2000$ CIFAR-100 training images at 224$\times$224 through the frozen model, and hook the output of the MLP post-expansion projection (\texttt{encoder.layers.encoder\_layer\_$k$.mlp[2]}) in each of the 12 encoder blocks.
We take the CLS-token activations as per-block ``channel'' vectors of dimension 768, then run the same Gaussian input-capture pipeline and use the predicted ImageNet-1k top-1 minus top-2 logit margin as the target signal.
Because the frozen ImageNet-1k ViT does not share the CIFAR-100 label space, this pilot is a predicted-margin representation-structure sanity check rather than task-aligned evidence comparable to the CIFAR-100 CNN experiments.
Table~\ref{tab:vit_pilot} reports per-block $\rho(I_X, I(T;Y))$ and ARI between local-axis and target-axis 3-means clusterings.
Mean $\rho = 0.010$ and mean ARI $= 0.044$ are within fluctuation of the CNN numbers in~\S\ref{sec:two_axes}, consistent with H1 beyond convolutional backbones while remaining a pilot rather than a decisive architecture-general proof.

\begin{table}[htbp]
\centering
\small
\begin{tabular}{lcc|lcc}
\toprule
Block & $\rho(I_X, I(T;Y))$ & ARI & Block & $\rho(I_X, I(T;Y))$ & ARI \\
\midrule
00 & $+0.134$ & $+0.192$ & 06 & $-0.060$ & $-0.045$ \\
01 & $-0.230$ & $-0.003$ & 07 & $-0.009$ & $-0.057$ \\
02 & $+0.383$ & $+0.011$ & 08 & $+0.147$ & $+0.031$ \\
03 & $+0.155$ & $-0.023$ & 09 & $-0.161$ & $+0.067$ \\
04 & $-0.020$ & $-0.040$ & 10 & $-0.093$ & $+0.148$ \\
05 & $+0.025$ & $+0.062$ & 11 & $-0.150$ & $+0.184$ \\
\midrule
Mean & $\mathbf{+0.010}$ & $\mathbf{+0.044}$ & & & \\
\bottomrule
\end{tabular}
\caption{\textbf{ViT-B/16 pilot: weak cross-axis alignment on a non-convolutional backbone.}
torchvision ViT-B/16, ImageNet-1k weights, CIFAR-100 calibration images (2000 images), and predicted ImageNet-1k top-1 minus top-2 margin as the target signal.
The overall mean $\rho$ and ARI are within fluctuation of the CNN suite despite considerably higher per-block variability, consistent with the two-axis picture beyond CNNs while still leaving full transformer validation to future work.}
\label{tab:vit_pilot}
\end{table}

\section{ImageNet-100 Recovery-Budget Sensitivity}
\label{app:imagenet_recovery}

This section is a limited robustness check, not part of the primary CIFAR-100 evidence base for H1--H4.
To test whether the weaker ImageNet-100 pruning curves are partly a recovery-budget artifact, we reran ResNet-50/ImageNet-100 with the same pruning protocol but increased the fine-tuning budget from 50 to 200 batches.
Recovery improves substantially at moderate sparsity: at 50\% sparsity, CAP-A rises from $21.5 \pm 1.3$\% to $45.9 \pm 1.2$\%, Taylor from $7.5 \pm 0.6$\% to $20.4 \pm 2.3$\%, and magnitude from $8.1 \pm 0.6$\% to $22.5 \pm 0.7$\%.
This confirms that some of the large-scale variance is a recovery-budget issue rather than a contradiction of the CIFAR-100 findings, while still leaving ImageNet-100 as robustness evidence rather than primary support.
Table~\ref{tab:imagenet_mb200} gives the 50-vs-200 batch numbers.

\begin{table}[htbp]
\centering
\caption{\textbf{ResNet-50/ImageNet-100 recovery-budget ablation.} Post-pruning accuracy under the fixed protocol with 50 vs.\ 200 fine-tuning batches.}
\label{tab:imagenet_mb200}
\small
\begin{tabular}{llccc}
\toprule
Sparsity & Method & MB=50 & MB=200 & $\Delta$ \\
\midrule
50\% & Taylor & 7.52 $\pm$ 0.57 & 20.35 $\pm$ 2.35 & +12.83 \\
50\% & CAP-A & 21.51 $\pm$ 1.27 & 45.94 $\pm$ 1.25 & +24.44 \\
50\% & CAP & 4.12 $\pm$ 0.85 & 10.94 $\pm$ 1.43 & +6.82 \\
50\% & Magnitude & 8.10 $\pm$ 0.58 & 22.52 $\pm$ 0.66 & +14.42 \\
90\% & Taylor & 1.19 $\pm$ 0.18 & 2.03 $\pm$ 0.14 & +0.84 \\
90\% & CAP-A & 1.31 $\pm$ 0.22 & 2.15 $\pm$ 0.42 & +0.84 \\
90\% & CAP & 1.20 $\pm$ 0.27 & 2.10 $\pm$ 0.49 & +0.90 \\
90\% & Magnitude & 1.27 $\pm$ 0.19 & 1.56 $\pm$ 0.07 & +0.29 \\
\bottomrule

\end{tabular}
\end{table}

\ifpaperincludechecklist
\clearpage
\input{checklist.tex}
\fi

\end{document}